\documentclass[sigconf]{acmart}

\AtBeginDocument{%
  }

\setcopyright{acmlicensed}
\copyrightyear{2018}
\acmYear{2018}
\acmDOI{XXXXXXX.XXXXXXX}
\acmConference[KDD '27]
{The 33rd ACM SIGKDD Conference on Knowledge Discovery and Data Mining}
{August, 2027}
{San Jose, CA, USA}
\acmISBN{978-1-4503-XXXX-X/2018/06}

\usepackage{multirow}
\usepackage{bm}
\usepackage[table]{xcolor}
\usepackage{cuted}
\usepackage{makecell}
\usepackage{adjustbox}

\usepackage{booktabs}
\usepackage{tabularx}
\usepackage{array}
\usepackage{amsmath}
\usepackage{algorithm}
\usepackage{algpseudocode}
\newcommand{\CI}[3]{%
  \makebox[7.2em][c]{%
    #1%
    \hspace{0.10em}%
    \raisebox{-0.45ex}[0pt][0pt]{%
      \scriptsize [#2,\,#3]%
    }%
  }%
}
\begin{document}

\title{EchoBridge: Long-Tail-Aware ECG--Echocardiography Text Alignment for Echocardiography-Derived Cardiac Findings}

\author{Xiaocheng Fang}
\affiliation{%
  \institution{School of Intelligence Science and Technology, Peking University}
  \city{Beijing}
  \country{China}
}
\additionalaffiliation{%
  \institution{State Key Laboratory of General Artificial Intelligence, Peking University}
  \city{Beijing}
  \country{China}
}
\email{fangxiaocheng26@stu.pku.edu.cn}

\author{Jieyi Cai}
\affiliation{%
  \institution{University of the Chinese Academy of Sciences}
  \city{Beijing}
  \country{China}
}
\email{caijieyi055@gmail.com}

\author{Guangkun Nie}
\authornotemark[1]
\affiliation{%
  \institution{School of Intelligence Science and Technology, Peking University}
  \city{Beijing}
  \country{China}
}
\email{nieguangkun@stu.pku.edu.cn}

\author{Haoyu Wang}
\affiliation{%
  \institution{University of the Chinese Academy of Sciences}
  \city{Beijing}
  \country{China}
}
\email{wanghaoyu252@mails.ucas.ac.cn}

\author{Jiarui Jin}
\authornotemark[1]
\affiliation{%
  \institution{School of Intelligence Science and Technology, Peking University}
  \city{Beijing}
  \country{China}
}
\email{jrjin25@stu.pku.edu.cn}

\author{Yujie Xiao}
\affiliation{%
  \institution{National Institute of Health Data Science, Peking University}
  \city{Beijing}
  \country{China}
}
\email{xiaoyujie@stu.pku.edu.cn}

\author{Bo Liu}
\authornotemark[1]
\affiliation{%
  \institution{School of Intelligence Science and Technology, Peking University}
  \city{Beijing}
  \country{China}
}
\email{liubo2022@stu.pku.edu.cn}

\author{Chenyang He}
\affiliation{%
  \institution{National Institute of Health Data Science, Peking University}
  \city{Beijing}
  \country{China}
}
\email{raoquan00@gmail.com}


\author{Qinghao Zhao}
\affiliation{%
\institution{Department of Cardiology, Peking University People’s Hospital}
\city{Beijing}
\country{China}
}
\email{qhzhao@pku.edu.cn}

\author{Gaofeng Cheng}
\affiliation{%
  \institution{University of the Chinese Academy of Sciences}
  \city{Beijing}
  \country{China}
}
\email{chenggaofeng@hccl.ioa.ac.cn}

\author{Hongyan Li}
\authornotemark[1]
\authornote{Corresponding authors.}
\affiliation{%
  \institution{School of Intelligence Science and Technology, Peking University}
  \city{Beijing}
  \country{China}
}
\email{leehy@pku.edu.cn}

\author{Shenda Hong}
\authornotemark[2]
\affiliation{%
  \institution{National Institute of Health Data Science, Peking University}
  \city{Beijing}
  \country{China}
}
\email{hongshenda@pku.edu.cn}

\renewcommand{\shortauthors}{Fang et al.}

\begin{abstract}
Standardized echocardiography conclusions provide meaningful supervision for learning ECG representations of echocardiography-derived cardiac findings. Global ECG--text alignment may entangle modality-specific factors, while long-tailed finding distributions provide sparse positive supervision for low-prevalence conditions. We propose EchoBridge with Complementary Shared--Private Projection (CSPP) and Adaptive Prototype Boundary Calibration (APBC). CSPP maps each modality into shared and auxiliary private projections, reduces directional redundancy via within-modality orthogonality, and bidirectionally aligns normalized shared projections. APBC organizes the shared hypersphere with class-specific prototypes, training-frequency-adaptive angular margins, and spherical Riesz repulsion. We evaluate EchoBridge on EchoNext-Mini and independent PKUPH and SHTMU cohorts under four protocols: prompt-based inference without downstream classifier training, in-domain frozen linear probing, target-domain cross-center frozen linear probing, and source-only cross-center transfer, supplemented by finding-specific analyses. EchoBridge improves classifier-free AUROC, AUPRC, and F1 over the strongest baselines by 7.88, 5.61, and 4.54 points, respectively, and achieves the highest point estimates across all in-domain and target-domain probing budgets and both source-only transfer cohorts. Finding-specific analyses show gains for most conditions, including several low-prevalence valvular findings. Code: \url{https://github.com/PKUDigitalHealth/EchoBridge}.
\end{abstract}

\ccsdesc[500]{Applied computing~Health informatics}
\ccsdesc[500]{Computing methodologies~Artificial intelligence}

\keywords{}


\maketitle

\begin{figure*}[t]
\centering
\includegraphics[width=0.96\linewidth]{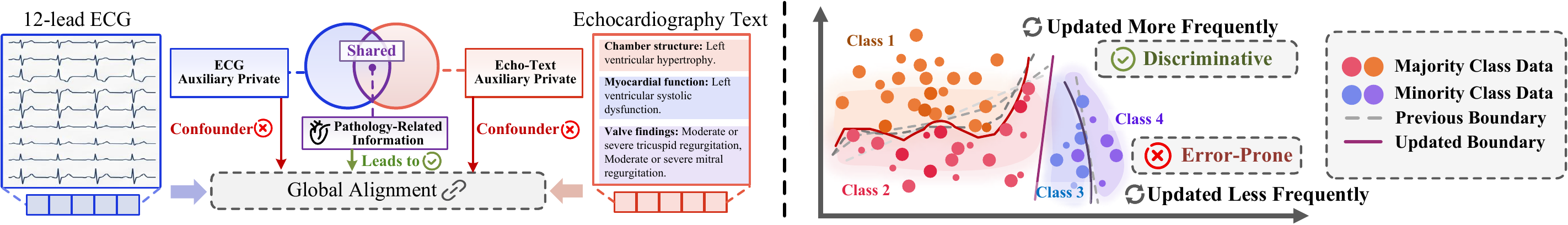}
\caption{Two challenges in ECG--echocardiography text alignment. Left: Global alignment entangles modality-specific factors and weakens clinically relevant cross-modal correspondence. Right: Prevalence imbalance provides fewer positive constraints and less reliable sample–prototype organization for low-prevalence findings.}
\label{Figure1}
\end{figure*}

\section{Introduction}
Echocardiography is a central imaging modality for evaluating cardiac structure, function, and valvular abnormalities~\cite{ghorbani2020deep,ouyang2020video,christensen2024vision}. Direct ECG–echocardiography learning from raw images or videos can be difficult to conduct at scale in retrospective multicenter studies because of storage, accessibility, and data-governance constraints~\cite{kaissis2020secure,vukadinovic2026comprehensive}. In routine practice, echocardiographic examinations are accompanied by diagnostic reports containing detailed findings and a concise conclusion that aggregates the principal structural, functional, and valvular abnormalities~\cite{chao2025echograph,kwak2025large}. We use these conclusion-level summaries as cross-modal supervision because they preserve clinically salient finding compositions while reducing measurement-, acquisition-, and template-specific variation. When paired with temporally matched ECG recordings, these summaries connect echocardiography-derived clinical semantics with cardiac electrical activity and provide a practical source of cross-modal supervision.

Recent AI--ECG studies show that ECGs encode echocardiography-confirmed abnormalities, including reduced ejection fraction and broader structural heart disease~\cite{attia2019screening,yao2021artificial,poterucha2025detecting}. MERL and ECG-CLIP use natural-language supervision for transferable representation learning~\cite{liu2024zero,zhou2025diagnosis}, while MERL-ECHO and Wearable-Echo-FM align ECGs with echocardiography reports for structural cardiac findings~\cite{wong2025contrastive,knight2026wearable}. Building on these methods, we learn a class-aware normalized space from paired ECGs and standardized conclusions, capturing paired-sample ECG--text alignment and organizing both modalities by predefined findings; evaluation covers prompt-based classifier-free inference for pretraining-seen findings, in-domain and target-domain cross-center frozen linear probing across label budgets, and source-only cross-center transfer.

Technically, aligning ECGs with standardized echocardiography conclusions poses two challenges. \textbf{1) Cross-modal representation interference.} ECG representations may encode continuous variation in rhythm, conduction, waveform morphology, and acquisition-related factors, whereas standardized conclusions provide compact compositional semantics for predefined structural, functional, and valvular findings. Global alignment may force heterogeneous ECG factors into a single conclusion-oriented embedding space, reducing the separation between finding-relevant and modality-dependent variation. \textbf{2) Long-tailed representation bias.} Echocardiography-derived findings often have highly imbalanced prevalence. Frequent findings contribute more positive samples and repeated optimization signals, potentially biasing the space toward head classes. Low-prevalence findings receive fewer positive constraints, reducing intra-class compactness and prototype separation and yielding less reliable decision boundaries and uneven representation quality across prevalence levels.

To address these challenges, we propose \textbf{EchoBridge}, an ECG--echocardiography text alignment framework for learning transferable ECG representations of echocardiography-derived findings. EchoBridge comprises two components. First, Complementary Shared--Private Projection maps ECG and text representations into alignment-oriented shared and auxiliary private projections. Cross-modal alignment and class-prototype supervision operate on the shared projections, while within-modality orthogonality reduces directional redundancy between branches and a symmetric contrastive objective aligns the $\ell_2$-normalized shared representations. Second, Adaptive Prototype Boundary Calibration organizes the shared hypersphere using class-specific prototypes for predefined findings. Training-frequency-adaptive angular margins strengthen positive sample--prototype constraints for low-prevalence findings, while spherical Riesz repulsion discourages prototype concentration. Together, these components integrate instance-level ECG--text correspondence with class-aware geometric organization under imbalanced finding distributions. We evaluate EchoBridge on EchoNext-Mini and two independent hospital cohorts from Peking University People's Hospital (PKUPH) and the Second Hospital of Tianjin Medical University (SHTMU) under four protocols: prompt-based classifier-free inference for pretraining-seen findings, in-domain frozen linear probing across multiple label budgets, target-domain cross-center linear probing, and source-only cross-center transfer, complemented by finding-specific analyses. The main contributions are as follows:

\begin{itemize}
\item We propose EchoBridge, an ECG--echocardiography text alignment framework for transferable ECG representations of echocardiography-derived findings under cross-modal interference and long-tailed distributions.

\item We develop Complementary Shared--Private Projection, which maps each modality into shared and auxiliary private projections, reduces directional redundancy via within-modality orthogonality, and bidirectionally aligns the normalized shared space.

\item We propose Adaptive Prototype Boundary Calibration, which structures the shared hypersphere with cross-modal class prototypes, training-frequency-adaptive angular margins, and spherical Riesz repulsion to improve class compactness and prototype separation under imbalance.

\item We evaluate EchoBridge on EchoNext-Mini and two independent hospital cohorts using prompt-based classifier-free inference, in-domain and target-domain cross-center frozen linear probing across label budgets, source-only cross-center transfer, and finding-specific analyses.
\end{itemize}

\begin{figure*}[t]
\centering
\includegraphics[width=\linewidth]{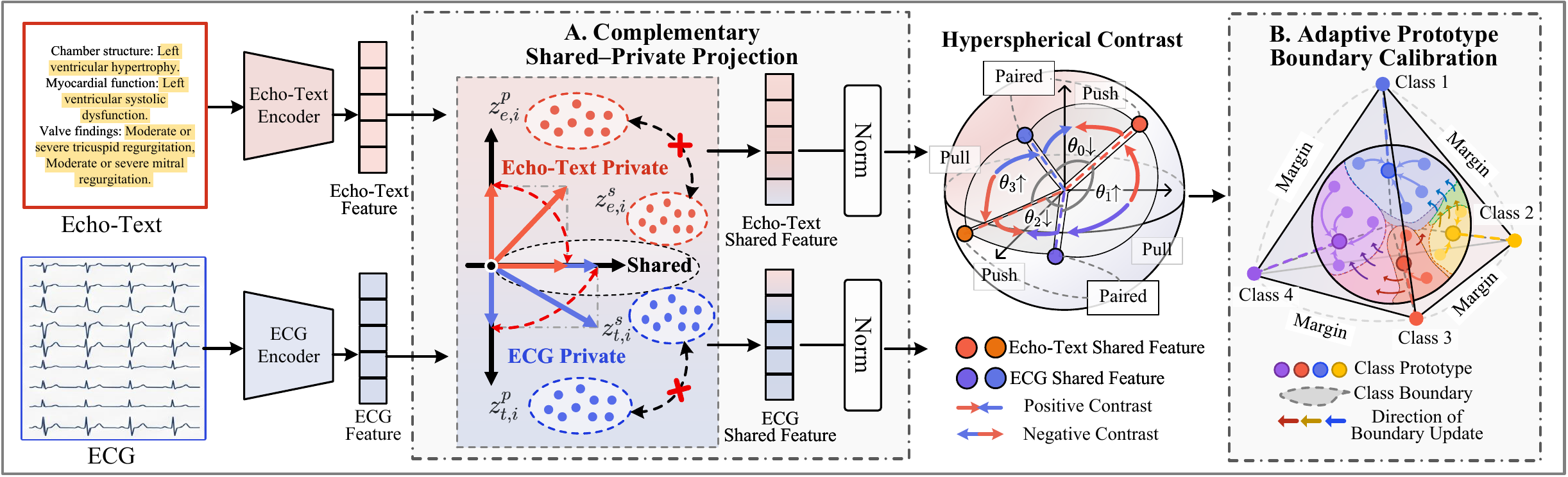}
\caption{Overview of EchoBridge. Given paired ECGs and echocardiography summaries, (A) Complementary Shared--Private Projection maps each modality into
alignment-oriented shared and auxiliary private branches, with
within-modality orthogonality reducing directional redundancy
between them. (B) Adaptive Prototype Boundary Calibration structures the shared space using class-specific prototypes, frequency-adaptive angular margins, and spherical Riesz repulsion.}

\label{Figure2}
\end{figure*}

\section{Methodology}
\subsection{Overview of EchoBridge}
EchoBridge learns transferable echocardiography-related ECG representations from paired 12-lead ECGs and standardized echocardiography conclusions. As shown in Figure~\ref{Figure2}, it combines Complementary Shared--Private Projection and Adaptive Prototype Boundary Calibration through normalized bidirectional alignment. Given an ECG $x_i$ and paired summary $t_i$, encoders $f_e(\cdot)$ and $f_t(\cdot)$ produce representations $h_{e,i}$ and $h_{t,i}$, which independent heads map into shared and auxiliary private projections. Within-modality orthogonality reduces directional redundancy between branches, while symmetric contrastive learning aligns the $\ell_2$-normalized shared representations by increasing paired-sample similarity over unpaired samples. Adaptive Prototype Boundary Calibration structures the shared space with class-specific prototypes for predefined cardiac findings. Training-frequency-adaptive angular margins strengthen positive alignment for low-prevalence findings, while spherical Riesz repulsion discourages prototype concentration and promotes pairwise separation on the unit hypersphere.

\subsection{Complementary Shared--Private Projection}
ECG waveforms encode rich continuous variation in rhythm, conduction, morphology, and acquisition conditions, whereas standardized echocardiography conclusions provide compact compositional semantics for predefined cardiac findings. Their global representations therefore contain factors with different relevance to cross-modal correspondence. Direct global alignment may entangle finding-relevant and modality-specific information within a single embedding space. We propose Complementary Shared--Private Projection, which maps each modality into alignment-oriented shared and auxiliary private projections. Cross-modal alignment and class-prototype supervision operate on the shared projections, while within-modality orthogonality reduces sample-wise directional redundancy between the two branches. The terms shared and private specify their optimization roles, while their detailed semantic contents remain unconstrained.

\paragraph{\bfseries Shared--Private Projection.}
Given an ECG $x_i$ and paired echocardiography conclusion $t_i$, the ECG and text encoders produce global representations:
\begin{equation}
h_{e,i}=f_e(x_i),
\qquad
h_{t,i}=f_t(t_i).
\end{equation}
Independent projection heads map each representation into shared and private projections:
\begin{equation}
z^s_{e,i}=\phi^s_e(h_{e,i}),
\qquad
z^p_{e,i}=\phi^p_e(h_{e,i}),
\end{equation}
\begin{equation}
z^s_{t,i}=\phi^s_t(h_{t,i}),
\qquad
z^p_{t,i}=\phi^p_t(h_{t,i}).
\end{equation}
Cross-modal alignment and class-prototype supervision operate on the shared projections $z^s_{e,i}$ and $z^s_{t,i}$. The independently parameterized private projections $z^p_{e,i}$ and $z^p_{t,i}$ provide auxiliary capacity without direct cross-modal alignment, allowing the shared branches to emphasize ECG--text correspondence.

To reduce sample-wise directional redundancy between the shared and private projections, we introduce a within-modality orthogonality objective:
\begin{equation}
\mathcal{L}_{\mathrm{orth}}
=
\frac{1}{N}
\sum_{i=1}^{N}
\left[
\left\langle
\bar{z}^s_{e,i},
\bar{z}^p_{e,i}
\right\rangle^2
+
\left\langle
\bar{z}^s_{t,i},
\bar{z}^p_{t,i}
\right\rangle^2
\right],
\end{equation}
where $N$ is the batch size, $\bar{z}=z/\|z\|_2$ denotes an $\ell_2$-normalized representation, and $\langle\cdot,\cdot\rangle$ denotes the inner product. Because the normalized inner product corresponds to cosine similarity, minimizing $\mathcal{L}_{\mathrm{orth}}$ drives the shared and private projections toward orthogonality within each modality. This geometric regularization reduces collinearity while leaving the semantic content of the shared and private projections unconstrained.

\paragraph{\bfseries Shared-Space Alignment.}
The shared ECG and text projections are normalized onto the unit hypersphere:
\begin{equation}
\hat{z}^s_{e,i}
=
\frac{z^s_{e,i}}{\|z^s_{e,i}\|_2},
\qquad
\hat{z}^s_{t,i}
=
\frac{z^s_{t,i}}{\|z^s_{t,i}\|_2}.
\end{equation}
For a mini-batch of $N$ paired samples, the cross-modal similarity matrix is defined as:
\begin{equation}
S_{ij}
=
\frac{
\left\langle
\hat{z}^s_{e,i},
\hat{z}^s_{t,j}
\right\rangle
}{\tau},
\end{equation}
where $\tau$ is the temperature parameter. Because both representations are $\ell_2$-normalized, their inner product equals cosine similarity.

We optimize bidirectional ECG--text correspondence using a symmetric contrastive objective:
\begin{equation}
\mathcal{L}_{\mathrm{align}}
=
\frac{1}{2N}
\sum_{i=1}^{N}
\left[
-\log
\frac{\exp(S_{ii})}
{\sum_{j=1}^{N}\exp(S_{ij})}
-
\log
\frac{\exp(S_{ii})}
{\sum_{j=1}^{N}\exp(S_{ji})}
\right].
\end{equation}
The first term retrieves the paired echocardiography conclusion from an ECG query, while the second performs reverse retrieval. This objective forms a common normalized space for paired ECG--text correspondence, which the following class-prototype objective organizes by echocardiography-derived findings.

\subsection{Adaptive Prototype Boundary Calibration}
Pairwise ECG--text alignment provides instance-level supervision, whereas fine-grained cardiac findings require explicit class-level organization. Long-tailed distributions may produce weak boundaries for infrequent classes, while unconstrained prototypes may cluster geometrically. We therefore propose Adaptive Prototype Boundary Calibration (APBC), combining frequency-adaptive angular margins calibrated by training-set label frequencies with spherical repulsion that separates normalized class prototypes.

We maintain a learnable prototype matrix:
\begin{equation}
P=[p_1,\ldots,p_C]^\top \in \mathbb{R}^{C\times d},
\end{equation}
where $C$ denotes the number of fine-grained cardiac findings and $d$ the shared-space dimension. Each prototype is normalized as:
\begin{equation}
\hat{p}_c=\frac{p_c}{\|p_c\|_2}.
\end{equation}
For normalized shared ECG and text representations $\hat{z}^{s}_{e,i}$ and $\hat{z}^{s}_{t,i}$, the prototype similarities are:
\begin{equation}
u^{e}_{i,c}
=
\left\langle \hat{z}^{s}_{e,i},\hat{p}_c \right\rangle,
\qquad
u^{t}_{i,c}
=
\left\langle \hat{z}^{s}_{t,i},\hat{p}_c \right\rangle.
\end{equation}
The corresponding logits are:
\begin{equation}
\ell^{e}_{i,c}=\gamma u^{e}_{i,c},
\qquad
\ell^{t}_{i,c}=\gamma u^{t}_{i,c},
\end{equation}
where $\gamma=\exp(s)$ is a learnable positive scale. Sharing $P$ across modalities provides a common category-level reference in the normalized space.

\paragraph{\bfseries{Frequency-Adaptive Angular Margin.}}
To address label-frequency imbalance, we assign each cardiac finding a class-specific angular margin based on its training-set positive rate. Let $r_c$ denote the positive rate of class $c$. Its margin is:
\begin{equation}
m_c
=
\mathrm{clip}
\left(
m_0
\cdot
\frac{\mathrm{median}(r)}
{r_c+\epsilon}
\cdot
\frac{1}{\kappa},
m_{\min},
m_{\max}
\right),
\end{equation}
where $m_0$ is the base margin, $\epsilon$ ensures numerical stability, and $m_{\min}$ and $m_{\max}$ bound the margin. The data-dependent normalization factor is:
\begin{equation}
\kappa
=
\frac{1}{C}
\sum_{c=1}^{C}
\frac{\mathrm{median}(r)}{r_c+\epsilon},
\end{equation}
which normalizes the mean pre-clipping margin to $m_0$. For each sample--class pair, the adjusted logits are:
\begin{equation}
\tilde{\ell}^{e}_{i,c}
=
\gamma
\cos
\left(
\theta^{e}_{i,c}
+
y_{i,c}m_c
\right),
\qquad
\tilde{\ell}^{t}_{i,c}
=
\gamma
\cos
\left(
\theta^{t}_{i,c}
+
y_{i,c}m_c
\right),
\end{equation}
where
$\theta^{e}_{i,c}=\arccos\left(u^{e}_{i,c}\right)$
and
$\theta^{t}_{i,c}=\arccos\left(u^{t}_{i,c}\right)$
denote the ECG--prototype and text--prototype angles, respectively,
and $y_{i,c}\in\{0,1\}$ indicates whether sample $i$ is positive
for class $c$. Thus, the margin applies only to positive sample--prototype pairs, leaving negative-pair logits unchanged. Lower-prevalence findings receive larger bounded margins, encouraging more discriminative representations around tail-class prototypes. The adjusted logits supervise the shared ECG and text branches:
\begin{equation}
\mathcal{L}_{\mathrm{proto}}
=
\mathrm{BCE}
\left(
\tilde{\ell}^{e},
y
\right)
+
\mathrm{BCE}
\left(
\tilde{\ell}^{t},
y
\right),
\end{equation}
where $y$ is the multi-label target matrix, and $\tilde{\ell}^{e}$ and $\tilde{\ell}^{t}$ are the adjusted ECG and text logits. This objective draws positive representations toward corresponding prototypes and suppresses responses to prototypes of negative findings, linking instance-level ECG--text alignment with class-level multi-label supervision.

\begin{figure}[t]
\centering
\includegraphics[width=\linewidth]{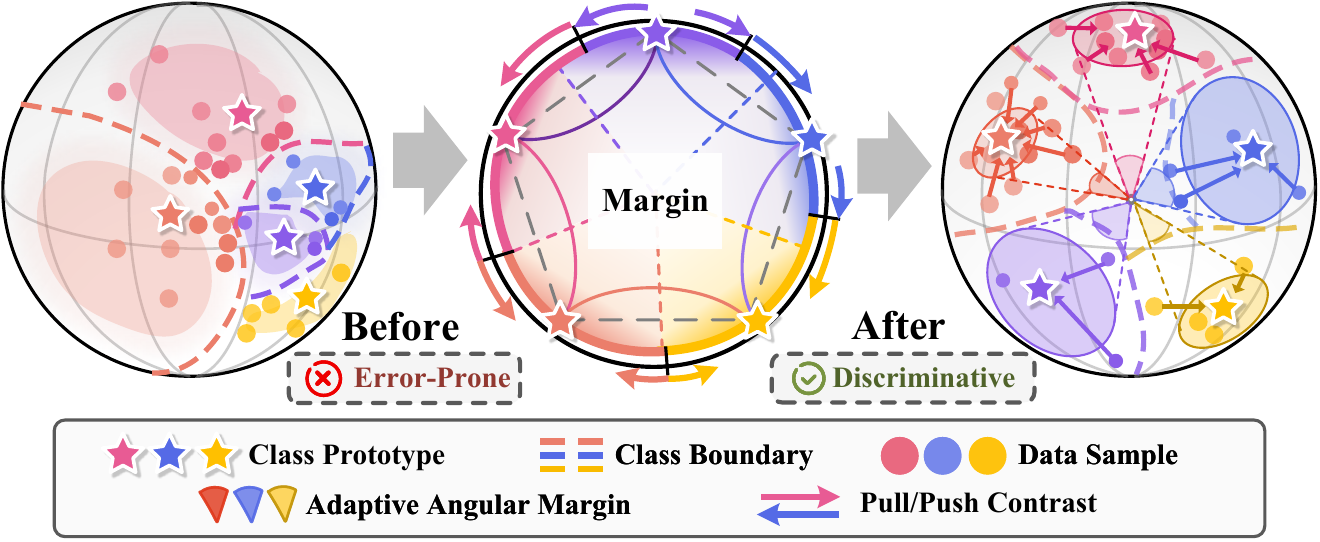}
\caption{Adaptive Prototype Boundary Calibration is designed to improve 
positive-class compactness and prototype separation on the 
hypersphere.}
\label{Figure3}
\end{figure}

\paragraph{\bfseries{Spherical Riesz Repulsion.}}
Prototype supervision constrains sample--prototype relationships, while class prototypes may still cluster on the hypersphere. We therefore apply a spherical Riesz repulsion loss to the normalized prototypes:
\begin{equation}
\mathcal{L}_{\mathrm{riesz}}
=
\frac{2}{C(C-1)}
\sum_{1\leq a<b\leq C}
\left\|
\hat{p}_a-\hat{p}_b
\right\|_2^{-q},
\qquad q>0.
\end{equation}
Because the loss increases as pairwise distance decreases, minimizing $\mathcal{L}_{\mathrm{riesz}}$ discourages prototype concentration and promotes separation on the unit hypersphere.

The Adaptive Prototype Boundary Calibration objective is:
\begin{equation}
\mathcal{L}_{\mathrm{apbc}}
=
\mathcal{L}_{\mathrm{proto}}
+
\lambda_r\mathcal{L}_{\mathrm{riesz}},
\end{equation}
where $\lambda_r$ controls the contribution of spherical prototype
repulsion.

\subsection{Overall Training Objective}
EchoBridge jointly optimizes Complementary Shared--Private Projection and Adaptive Prototype Boundary Calibration with the objective:
\begin{equation}
\mathcal{L}
=
\mathcal{L}_{\mathrm{align}}
+
\mathcal{L}_{\mathrm{orth}}
+
\mathcal{L}_{\mathrm{apbc}}.
\end{equation}

\section{Experiments}
\subsection{Datasets and Splits}
\begin{figure}[t]
\centering
\includegraphics[width=\linewidth]{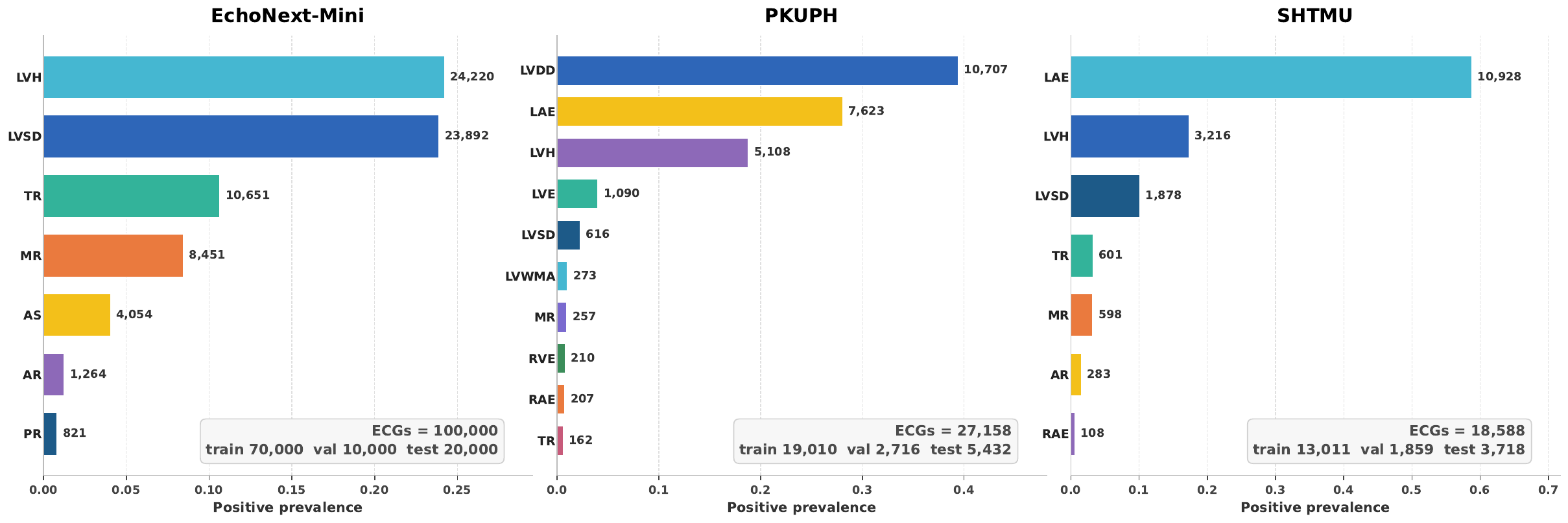}
\caption{Dataset statistics and label distributions across EchoNext-Mini, PKUPH, and SHTMU. Each panel also reports the ECG cohort size and the train/validation/test split.}
\label{Figure4}
\end{figure}

We evaluate EchoBridge on the open-source EchoNext-Mini~\cite{hughes2026echonext} and two private real-world cohorts from Peking University People's Hospital (PKUPH) and the Second Hospital of Tianjin Medical University (SHTMU). Each sample pairs an ECG with echocardiography-derived findings
from a temporally matched examination under a dataset-specific
matching window. We partition each dataset into patient-disjoint training, validation, and test sets using a 7:1:2 ratio to reduce information leakage. All datasets exhibit long-tailed label distributions; Figure~\ref{Figure4} reports their statistics, label distributions, and split sizes.

\subsection{Data Preprocessing}

\paragraph{\bfseries ECG Signal Processing.}
We applied record-level quality control and excluded unreadable ECGs, records with substantial missingness, and samples that could not be reliably linked to patient identifiers or labels. All signals were resampled to 500~Hz by linear interpolation and processed using a fixed denoising pipeline comprising a 0.5-Hz high-pass filter, a second-order Butterworth low-pass filter with a 50-Hz cutoff, and a 50/60-Hz notch filter. Recordings were standardized to 10-second segments, with longer recordings divided into consecutive temporal windows. Each segment was Z-score normalized, and missing leads or signal values were zero-filled to preserve a consistent input shape.

\paragraph{\bfseries Echocardiography Conclusion-Style Text Construction.}
Clinical echocardiography reports typically contain detailed findings and a concise conclusion or impression. Findings may include quantitative measurements, image-quality descriptions, and contextual observations with variable relevance to predefined cardiac findings, limiting their consistency as supervision for finding-specific representation learning. We therefore use conclusion-level semantics. For each examination, we deterministically verbalize the structured multi-label finding vector into a standardized conclusion-style summary organized by chamber structure, myocardial function, and valvular abnormalities. This preserves multi-label co-occurrence, mirrors the concise compositional form of clinical conclusions, and controls unrelated lexical, formatting, and template variation. As an informal quality check, five senior cardiologists inspected a 
random subset of the generated conclusion-style summaries and provided qualitative feedback that the sampled summaries were broadly clinically plausible and semantically consistent with the corresponding structured findings. This review was not designed as a formal annotation or inter-rater agreement study. The resulting text supports ECG--text alignment and class-specific prototype learning under controlled conclusion-level semantics.

\subsection{Baselines and Implementation Details}
We compare EchoBridge with representative ECG-only self-supervised and ECG--text pretraining methods. Prompt-based classifier-free inference uses ECG--text dual encoders with fixed definition-only textual prototypes for pretraining-seen findings. In all linear-probing experiments, the pretrained ECG representation extractor is frozen, and an identical linear multi-label classifier is trained with 1\%, 10\%, or 100\% of the available labels. All methods share patient-level partitions, ECG preprocessing, label subsets, metrics, and validation-based threshold selection. We use official implementations when available; otherwise, we reproduce methods under matched backbone, optimization, and training-budget settings where applicable. Appendix~\ref{sec:supervised_comparison} further compares EchoBridge with task-specific ResNet-18 classifiers trained end-to-end on EchoNext-Mini.

EchoBridge was implemented in PyTorch and trained on one NVIDIA A100 GPU using a one-dimensional ResNet-18 ECG encoder and MedCPT-Article-Encoder text encoder~\cite{jin2023medcpt}. AdamW optimization uses a learning rate of $1\times10^{-5}$, weight decay of $1\times10^{-8}$, batch size 64, and up to 15 epochs. Further architecture and optimization details are provided in the appendix.

\begin{table}[t]
\centering
\small
\setlength{\tabcolsep}{3pt}
\renewcommand{\arraystretch}{1.05}
\caption{
Prompt-based classifier-free inference performance on EchoNext-Mini for pretraining-seen cardiac findings.
}
\label{tab:EchoNext-Mini}
\resizebox{\columnwidth}{!}{%
\begin{tabular}{@{}llccc@{}}
\toprule
\textbf{Methods}
& \textbf{Ref.}
& \textbf{AUROC}
& \textbf{AUPRC}
& \textbf{F1} \\
\midrule
CLIP~\cite{radford2021learning}
& ICML'21
& \CI{62.57}{61.70}{63.41}
& \CI{17.03}{16.42}{17.82}
& \CI{24.13}{23.49}{25.17} \\
SigLIP~\cite{zhai2023sigmoid}
& ICCV'23
& \CI{65.07}{64.23}{65.91}
& \CI{17.35}{16.73}{18.14}
& \CI{25.00}{24.36}{26.04} \\
PCME++~\cite{chun2023improved}
& ICLR'24
& \CI{62.02}{61.14}{62.87}
& \CI{17.59}{16.95}{18.38}
& \CI{23.07}{22.39}{24.06} \\
MERL-ECHO~\cite{wong2025contrastive}
& medRxiv'25
& \CI{64.02}{63.17}{64.86}
& \CI{18.63}{17.98}{19.48}
& \CI{23.92}{23.26}{24.87} \\
ECG-CLIP~\cite{zhou2025diagnosis}
& npj DM'25
& \CI{60.47}{59.55}{61.36}
& \CI{15.47}{14.87}{16.22}
& \CI{22.91}{22.28}{23.89} \\
D-BETA~\cite{hung2025boosting}
& ICML'25
& \CI{63.99}{63.15}{64.83}
& \CI{20.12}{19.46}{20.96}
& \CI{\underline{27.29}}{26.58}{28.36} \\
SGERA~\cite{chen2026sgera}
& ICML'26
& \CI{\underline{67.85}}{67.03}{68.67}
& \CI{\underline{21.18}}{20.51}{22.03}
& \CI{26.95}{26.26}{27.99} \\
\midrule
\textbf{EchoBridge}
& \textbf{Ours}
& \CI{\textbf{75.73}}{75.00}{76.54}
& \CI{\textbf{26.79}}{25.95}{27.98}
& \CI{\textbf{31.83}}{31.15}{33.30} \\
\bottomrule
\end{tabular}%
}
\end{table}

\subsection{Evaluation Protocols and Metrics}
We evaluate pretrained ECG representations under four protocols. Prompt-based classifier-free inference predicts pretraining-seen findings by comparing frozen ECG representations with fixed definition-only textual prototypes. In-domain frozen linear probing trains a linear multi-label classifier on EchoNext-Mini using 1\%, 10\%, or 100\% of the available labels. Target-domain cross-center probing freezes the EchoNext-Mini-pretrained extractor and trains a new linear classifier on PKUPH or SHTMU with the same label ratios. Source-only transfer trains the classifier and selects thresholds exclusively on EchoNext-Mini, then evaluates findings shared with each external cohort. All experiments use patient-disjoint test sets and report macro-averaged AUROC, AUPRC, and F1 as percentage values. Class-specific thresholds maximize validation-set F1 and remain fixed for testing; source-only transfer retains EchoNext-Mini thresholds. We estimate 95\% confidence intervals using 1,000 non-parametric bootstrap repetitions. Appendix~\ref{app:evaluation_protocols} provides detailed protocols and evaluation procedures.

\begin{table*}[t]
\centering
\small
\setlength{\tabcolsep}{0.3pt}
\renewcommand{\arraystretch}{1.05}
\caption{
Frozen linear-probing performance on EchoNext-Mini using 1\%, 10\%, and 100\% of the available training labels. Each entry reports the macro-averaged metric value with its 95\% bootstrap confidence interval in brackets.
}
\label{tab:echonext_mini_ratio}

\begin{adjustbox}{max width=\textwidth,center}
\begin{tabular}{@{}llccc|ccc|ccc@{}}
\toprule
\multirow{2}{*}{\textbf{Methods}}
&
\multirow{2}{*}{\textbf{Ref.}}
&
\multicolumn{3}{c}{\textbf{1\% Linear Probing}}
&
\multicolumn{3}{c}{\textbf{10\% Linear Probing}}
&
\multicolumn{3}{c}{\textbf{100\% Linear Probing}}
\\
\cmidrule(lr){3-5}
\cmidrule(lr){6-8}
\cmidrule(lr){9-11}
&
&
\textbf{AUROC}
&
\textbf{AUPRC}
&
\textbf{F1}
&
\textbf{AUROC}
&
\textbf{AUPRC}
&
\textbf{F1}
&
\textbf{AUROC}
&
\textbf{AUPRC}
&
\textbf{F1}
\\
\midrule

\multicolumn{11}{@{}l}{\textbf{ECG-only Self-Supervised Learning}} \\

SimCLR~\cite{chen2020simple}
&
ICML'20
&
\CI{60.38}{59.42}{61.39}
&
\CI{13.78}{13.20}{14.58}
&
\CI{20.55}{19.85}{21.62}
&
\CI{62.19}{61.33}{63.10}
&
\CI{16.81}{16.15}{17.75}
&
\CI{23.17}{22.43}{24.37}
&
\CI{68.70}{67.92}{69.53}
&
\CI{19.96}{19.10}{21.06}
&
\CI{25.65}{24.85}{27.02}
\\

ST-MEM~\cite{na2024guiding}
&
ICLR'24
&
\CI{62.44}{61.41}{63.40}
&
\CI{17.09}{16.47}{17.86}
&
\CI{22.74}{22.00}{23.78}
&
\CI{68.42}{67.49}{69.28}
&
\CI{19.92}{19.22}{20.83}
&
\CI{25.92}{25.14}{27.09}
&
\CI{71.16}{70.31}{71.94}
&
\CI{21.72}{20.82}{22.79}
&
\CI{27.57}{26.73}{28.91}
\\

HeartLang~\cite{jin2025reading}
&
ICLR'25
&
\CI{60.15}{59.17}{61.18}
&
\CI{16.98}{16.39}{17.79}
&
\CI{21.86}{21.15}{22.94}
&
\CI{70.78}{69.90}{71.71}
&
\CI{22.58}{21.91}{23.53}
&
\CI{27.41}{26.66}{28.62}
&
\CI{74.59}{73.79}{75.44}
&
\CI{25.37}{24.50}{26.48}
&
\CI{30.69}{29.88}{32.07}
\\

\midrule
\multicolumn{11}{@{}l}{\textbf{ECG--Text Pretraining}} \\

CLIP~\cite{radford2021learning}
&
ICML'21
&
\CI{60.11}{59.06}{61.09}
&
\CI{14.30}{13.67}{15.08}
&
\CI{22.12}{21.37}{23.17}
&
\CI{64.99}{64.04}{65.87}
&
\CI{18.97}{18.26}{19.89}
&
\CI{23.65}{22.86}{24.83}
&
\CI{70.46}{69.59}{71.26}
&
\CI{22.14}{21.23}{23.22}
&
\CI{26.86}{26.01}{28.21}
\\

SigLIP~\cite{zhai2023sigmoid}
&
ICCV'23
&
\CI{61.59}{60.60}{62.53}
&
\CI{15.35}{14.76}{16.11}
&
\CI{22.63}{21.92}{23.66}
&
\CI{66.24}{65.35}{67.08}
&
\CI{19.43}{18.76}{20.33}
&
\CI{23.91}{23.16}{25.07}
&
\CI{70.97}{70.16}{71.73}
&
\CI{21.87}{21.00}{22.93}
&
\CI{27.11}{26.30}{28.44}
\\

PCME++~\cite{chun2023improved}
&
ICLR'24
&
\CI{57.08}{56.04}{58.12}
&
\CI{12.74}{12.12}{13.56}
&
\CI{20.04}{19.30}{21.13}
&
\CI{63.04}{62.10}{63.98}
&
\CI{17.62}{16.92}{18.58}
&
\CI{22.77}{21.99}{23.99}
&
\CI{69.98}{69.12}{70.84}
&
\CI{20.95}{20.05}{22.07}
&
\CI{26.09}{25.25}{27.48}
\\

MERL-ECHO~\cite{wong2025contrastive}
&
medRxiv'25
&
\CI{61.90}{60.95}{62.87}
&
\CI{16.04}{15.47}{16.81}
&
\CI{22.80}{22.11}{23.84}
&
\CI{66.36}{65.51}{67.23}
&
\CI{19.40}{18.75}{20.31}
&
\CI{25.01}{24.28}{26.18}
&
\CI{71.90}{71.13}{72.69}
&
\CI{23.32}{22.47}{24.39}
&
\CI{29.16}{28.37}{30.50}
\\

ECG-CLIP~\cite{zhou2025diagnosis}
&
npj DM'25
&
\CI{61.56}{60.54}{62.58}
&
\CI{17.71}{17.10}{18.51}
&
\CI{23.10}{22.37}{24.17}
&
\CI{69.57}{68.65}{70.49}
&
\CI{21.80}{21.11}{22.74}
&
\CI{27.01}{26.24}{28.21}
&
\CI{73.89}{73.05}{74.73}
&
\CI{24.27}{23.38}{25.37}
&
\CI{29.86}{29.03}{31.23}
\\

D-BETA~\cite{hung2025boosting}
&
ICML'25
&
\CI{\underline{68.99}}{67.99}{69.94}
&
\CI{23.02}{22.42}{23.78}
&
\CI{\underline{28.08}}{27.36}{29.11}
&
\CI{\underline{75.93}}{75.03}{76.78}
&
\CI{26.84}{26.16}{27.74}
&
\CI{30.41}{29.65}{31.57}
&
\CI{76.43}{75.61}{77.20}
&
\CI{29.13}{28.25}{30.19}
&
\CI{32.82}{32.00}{34.15}
\\

SGERA~\cite{chen2026sgera}
&
ICML'26
&
\CI{68.76}{67.79}{69.76}
&
\CI{\underline{24.49}}{23.91}{25.28}
&
\CI{27.73}{27.03}{28.79}
&
\CI{74.97}{74.10}{75.87}
&
\CI{\underline{27.69}}{27.03}{28.62}
&
\CI{\underline{31.74}}{31.00}{32.93}
&
\CI{\underline{77.19}}{76.40}{78.01}
&
\CI{\underline{30.25}}{29.39}{31.34}
&
\CI{\underline{33.57}}{32.77}{34.93}
\\

\midrule

\textbf{EchoBridge}
&
\textbf{Ours}
&
\CI{\textbf{72.77}}{71.84}{73.79}
&
\CI{\textbf{26.53}}{25.67}{27.56}
&
\CI{\textbf{31.32}}{30.49}{32.59}
&
\CI{\textbf{76.94}}{76.10}{77.84}
&
\CI{\textbf{30.45}}{29.55}{31.52}
&
\CI{\textbf{34.13}}{33.18}{35.52}
&
\CI{\textbf{78.79}}{77.98}{79.55}
&
\CI{\textbf{32.66}}{31.65}{33.88}
&
\CI{\textbf{35.82}}{35.02}{37.36}
\\

\bottomrule
\end{tabular}
\end{adjustbox}
\end{table*}

\begin{table*}[t]
\centering
\small
\setlength{\tabcolsep}{0.3pt}
\renewcommand{\arraystretch}{1.05}
\caption{
Target-domain cross-center frozen linear-probing performance on PKUPH using 1\%, 10\%, and 100\% of the available target-cohort training labels. All ECG representation extractors are pretrained on EchoNext-Mini and frozen; only a linear multi-label classifier is trained on PKUPH.
}
\label{tab:pkuph_ratio}

\begin{adjustbox}{max width=\textwidth,center}
\begin{tabular}{@{}llccc|ccc|ccc@{}}
\toprule
\multirow{2}{*}{\textbf{Methods}}
&
\multirow{2}{*}{\textbf{Ref.}}
&
\multicolumn{3}{c}{\textbf{1\% Linear Probing}}
&
\multicolumn{3}{c}{\textbf{10\% Linear Probing}}
&
\multicolumn{3}{c}{\textbf{100\% Linear Probing}}
\\
\cmidrule(lr){3-5}
\cmidrule(lr){6-8}
\cmidrule(lr){9-11}
&
&
\textbf{AUROC}
&
\textbf{AUPRC}
&
\textbf{F1}
&
\textbf{AUROC}
&
\textbf{AUPRC}
&
\textbf{F1}
&
\textbf{AUROC}
&
\textbf{AUPRC}
&
\textbf{F1}
\\
\midrule

\multicolumn{11}{@{}l}{\textbf{ECG-only Self-Supervised Learning}} \\

SimCLR~\cite{chen2020simple}
&
ICML'20
&
\CI{55.34}{52.99}{57.57}
&
\CI{11.92}{11.56}{12.44}
&
\CI{17.92}{17.43}{18.85}
&
\CI{59.47}{56.94}{61.89}
&
\CI{12.92}{12.52}{13.47}
&
\CI{18.29}{17.87}{19.35}
&
\CI{72.61}{70.40}{74.89}
&
\CI{18.51}{17.40}{20.57}
&
\CI{27.30}{25.90}{30.01}
\\

ST-MEM~\cite{na2024guiding} & ICLR'24 & \CI{58.12}{55.73}{60.46} & \CI{12.54}{12.03}{13.26} & \CI{18.73}{18.06}{20.42} & \CI{69.84}{67.41}{72.18} & \CI{16.48}{15.57}{18.07} & \CI{23.46}{22.31}{26.05} & \CI{76.84}{74.72}{78.93} & \CI{22.91}{21.34}{24.76} & \CI{30.92}{29.36}{33.18} \\
HeartLang~\cite{jin2025reading}
&
ICLR'25
&
\CI{56.29}{53.92}{58.64}
&
\CI{11.99}{11.64}{12.67}
&
\CI{18.18}{17.64}{20.38}
&
\CI{71.18}{68.85}{73.40}
&
\CI{17.25}{16.45}{18.95}
&
\CI{24.50}{23.51}{27.64}
&
\CI{76.95}{74.88}{78.86}
&
\CI{22.90}{21.28}{25.94}
&
\CI{31.31}{30.80}{36.02}
\\

\midrule
\multicolumn{11}{@{}l}{\textbf{ECG--Text Pretraining}} \\

CLIP~\cite{radford2021learning}
&
ICML'21
&
\CI{65.44}{62.74}{68.11}
&
\CI{16.27}{14.93}{18.52}
&
\CI{25.07}{23.38}{28.33}
&
\CI{68.92}{66.26}{71.35}
&
\CI{18.24}{16.78}{20.61}
&
\CI{26.11}{24.68}{29.63}
&
\CI{74.53}{72.14}{76.92}
&
\CI{21.25}{19.64}{23.73}
&
\CI{28.87}{27.38}{32.18}
\\

SigLIP~\cite{zhai2023sigmoid}
&
ICCV'23
&
\CI{68.51}{66.26}{71.03}
&
\CI{17.70}{16.29}{20.13}
&
\CI{26.65}{25.07}{30.34}
&
\CI{68.80}{66.02}{71.56}
&
\CI{18.97}{17.45}{21.49}
&
\CI{27.54}{25.98}{31.13}
&
\CI{73.01}{70.68}{75.49}
&
\CI{20.18}{18.65}{22.67}
&
\CI{28.56}{26.67}{32.17}
\\

PCME++~\cite{chun2023improved}
&
ICLR'24
&
\CI{63.29}{60.39}{66.07}
&
\CI{14.61}{13.77}{16.17}
&
\CI{22.55}{21.42}{25.20}
&
\CI{66.98}{64.51}{69.54}
&
\CI{15.73}{14.90}{17.33}
&
\CI{22.90}{21.82}{25.44}
&
\CI{75.06}{73.17}{76.86}
&
\CI{18.77}{17.63}{20.75}
&
\CI{26.39}{24.89}{29.49}
\\

MERL-ECHO~\cite{wong2025contrastive}
&
medRxiv'25
&
\CI{67.79}{64.92}{70.24}
&
\CI{15.83}{15.02}{17.37}
&
\CI{24.21}{22.90}{26.96}
&
\CI{71.55}{68.85}{73.80}
&
\CI{17.54}{16.71}{19.17}
&
\CI{25.08}{23.87}{28.18}
&
\CI{77.88}{76.88}{80.71}
&
\CI{21.24}{20.11}{23.29}
&
\CI{29.90}{28.54}{32.94}
\\

ECG-CLIP~\cite{zhou2025diagnosis}
&
npj DM'25
&
\CI{68.47}{65.83}{70.94}
&
\CI{18.34}{16.72}{20.05}
&
\CI{\underline{28.05}}{26.41}{29.20}
&
\CI{72.66}{70.13}{74.86}
&
\CI{19.72}{18.03}{21.24}
&
\CI{\underline{29.18}}{27.38}{30.42}
&
\CI{\underline{78.02}}{76.01}{79.12}
&
\CI{23.62}{22.01}{25.12}
&
\CI{31.76}{30.18}{33.16}
\\

D-BETA~\cite{hung2025boosting}
&
ICML'25
&
\CI{69.63}{67.06}{71.84}
&
\CI{\underline{18.91}}{17.23}{20.19}
&
\CI{27.42}{25.68}{29.08}
&
\CI{\underline{73.58}}{71.12}{75.18}
&
\CI{20.10}{18.39}{21.52}
&
\CI{28.76}{26.91}{30.31}
&
\CI{77.48}{75.41}{79.03}
&
\CI{23.84}{22.21}{25.18}
&
\CI{\underline{32.06}}{30.45}{33.22}
\\

SGERA~\cite{chen2026sgera}
&
ICML'26
&
\CI{\underline{70.91}}{68.42}{72.47}
&
\CI{18.62}{16.91}{20.10}
&
\CI{27.88}{26.03}{29.16}
&
\CI{73.21}{70.75}{75.12}
&
\CI{\underline{20.51}}{18.72}{21.61}
&
\CI{28.96}{27.02}{30.37}
&
\CI{77.66}{75.62}{79.17}
&
\CI{\underline{24.11}}{22.47}{25.29}
&
\CI{31.84}{30.26}{33.19}
\\

\midrule

\textbf{EchoBridge}
&
\textbf{Ours}
&
\CI{\textbf{72.68}}{70.26}{75.00}
&
\CI{\textbf{20.32}}{18.31}{23.88}
&
\CI{\textbf{29.31}}{27.44}{33.90}
&
\CI{\textbf{75.42}}{72.92}{77.72}
&
\CI{\textbf{21.79}}{19.82}{25.28}
&
\CI{\textbf{30.60}}{28.55}{35.02}
&
\CI{\textbf{79.48}}{77.32}{81.46}
&
\CI{\textbf{25.43}}{23.70}{28.62}
&
\CI{\textbf{33.40}}{31.86}{37.62}
\\

\bottomrule
\end{tabular}
\end{adjustbox}
\end{table*}

\section{Results}
\subsection{Prompt-Based Classifier-Free Inference}
Table~\ref{tab:EchoNext-Mini} evaluates whether the frozen shared space supports classifier-free prediction using fixed definition-only textual prototypes. EchoBridge achieves 75.73 AUROC, 26.79 AUPRC, and 31.83 F1, exceeding the strongest competing method for each metric by 7.88, 5.61, and 4.54 percentage points, respectively. These gains indicate improved score ranking and positive-finding discrimination under imbalanced multi-label evaluation. Because all evaluated findings are incorporated during pretraining through standardized conclusion-style summaries and class-specific prototype supervision, this protocol measures the accessibility of pretraining-seen finding semantics through textual prototypes. The results support stronger correspondence between ECG representations and predefined echocardiography-derived findings.

\subsection{In-Domain Frozen Linear Probing under Different Label Ratios}
Table~\ref{tab:echonext_mini_ratio} evaluates frozen ECG representations on EchoNext-Mini using linear multi-label classifiers trained with 1\%, 10\%, or 100\% of the available labels. EchoBridge achieves the highest AUROC, AUPRC, and F1 among the evaluated ECG-only and ECG--echocardiography text methods at every budget. With 1\% labels, it obtains 72.77 AUROC, 26.53 AUPRC, and 31.32 F1, exceeding the strongest baseline for each metric by 3.78, 2.04, and 3.24 points, respectively. These gains indicate that finding-related information remains linearly accessible under limited supervision. At 10\%, the corresponding improvements are 1.01, 2.76, and 2.39 points; at 100\%, they are 1.60, 2.41, and 2.25 points. Consistent AUPRC and F1 gains indicate improved positive-finding discrimination under class imbalance. ECG--summary alignment with class-aware prototype supervision therefore yields frozen representations that remain linearly accessible across in-domain label budgets.

\begin{table*}[t]
\centering
\small
\setlength{\tabcolsep}{0.3pt}
\renewcommand{\arraystretch}{1.05}
\caption{
Target-domain cross-center frozen linear-probing performance on SHTMU using 1\%, 10\%, and 100\% of the available target-cohort training labels. All ECG representation extractors are pretrained on EchoNext-Mini and frozen; only a linear multi-label classifier is trained on SHTMU. 
}
\label{tab:shtmu_ratio}

\begin{adjustbox}{max width=\textwidth,center}
\begin{tabular}{@{}llccc|ccc|ccc@{}}
\toprule
\multirow{2}{*}{\textbf{Methods}}
&
\multirow{2}{*}{\textbf{Ref.}}
&
\multicolumn{3}{c}{\textbf{1\% Linear Probing}}
&
\multicolumn{3}{c}{\textbf{10\% Linear Probing}}
&
\multicolumn{3}{c}{\textbf{100\% Linear Probing}}
\\
\cmidrule(lr){3-5}
\cmidrule(lr){6-8}
\cmidrule(lr){9-11}
&
&
\textbf{AUROC}
&
\textbf{AUPRC}
&
\textbf{F1}
&
\textbf{AUROC}
&
\textbf{AUPRC}
&
\textbf{F1}
&
\textbf{AUROC}
&
\textbf{AUPRC}
&
\textbf{F1}
\\
\midrule

\multicolumn{11}{@{}l}{\textbf{ECG-only Self-Supervised Learning}} \\

SimCLR~\cite{chen2020simple}
&
ICML'20
&
\CI{56.20}{53.85}{58.46}
&
\CI{15.95}{15.46}{16.78}
&
\CI{22.93}{22.20}{24.50}
&
\CI{63.43}{61.00}{65.61}
&
\CI{18.85}{18.20}{19.96}
&
\CI{25.05}{24.57}{26.96}
&
\CI{69.91}{67.50}{72.05}
&
\CI{23.48}{22.62}{24.99}
&
\CI{29.33}{28.42}{31.36}
\\
ST-MEM~\cite{na2024guiding} & ICLR'24 & \CI{57.02}{54.61}{59.31} & \CI{16.71}{15.96}{17.83} & \CI{23.86}{22.98}{25.96} & \CI{65.91}{63.53}{68.17} & \CI{21.38}{20.46}{22.86} & \CI{27.44}{26.52}{29.72} & \CI{73.88}{71.81}{75.81} & \CI{25.92}{24.88}{27.65} & \CI{31.68}{30.61}{32.87} \\
HeartLang~\cite{jin2025reading}
&
ICLR'25
&
\CI{55.41}{52.94}{57.59}
&
\CI{16.38}{15.62}{17.49}
&
\CI{23.20}{22.46}{25.09}
&
\CI{67.48}{65.52}{69.65}
&
\CI{\underline{23.07}}{22.10}{24.54}
&
\CI{28.56}{27.88}{30.34}
&
\CI{73.39}{71.60}{75.29}
&
\CI{25.12}{24.15}{26.89}
&
\CI{31.10}{30.28}{33.56}
\\

\midrule
\multicolumn{11}{@{}l}{\textbf{ECG--Text Pretraining}} \\

CLIP~\cite{radford2021learning}
&
ICML'21
&
\CI{59.06}{56.56}{61.72}
&
\CI{17.95}{17.09}{19.80}
&
\CI{25.28}{24.09}{27.98}
&
\CI{64.25}{61.72}{66.79}
&
\CI{19.93}{19.10}{21.65}
&
\CI{26.66}{25.67}{29.20}
&
\CI{69.11}{66.98}{71.26}
&
\CI{24.04}{22.88}{26.12}
&
\CI{29.97}{28.57}{32.79}
\\

SigLIP~\cite{zhai2023sigmoid}
&
ICCV'23
&
\CI{60.55}{58.06}{63.07}
&
\CI{18.19}{17.25}{20.23}
&
\CI{26.22}{24.71}{28.97}
&
\CI{63.23}{60.78}{65.62}
&
\CI{19.60}{18.74}{21.18}
&
\CI{26.80}{25.29}{29.60}
&
\CI{66.47}{64.22}{68.74}
&
\CI{22.11}{20.96}{24.26}
&
\CI{28.58}{26.93}{31.81}
\\

PCME++~\cite{chun2023improved}
&
ICLR'24
&
\CI{60.01}{57.56}{62.44}
&
\CI{17.28}{16.57}{18.36}
&
\CI{24.47}{23.66}{26.50}
&
\CI{61.80}{59.37}{64.32}
&
\CI{18.33}{17.61}{19.51}
&
\CI{25.18}{24.35}{27.57}
&
\CI{68.58}{66.46}{70.63}
&
\CI{21.90}{21.00}{23.37}
&
\CI{27.97}{27.00}{30.52}
\\

MERL-ECHO~\cite{wong2025contrastive}
&
medRxiv'25
&
\CI{64.11}{61.47}{66.72}
&
\CI{18.58}{17.75}{20.03}
&
\CI{26.34}{25.28}{28.80}
&
\CI{66.56}{63.89}{69.09}
&
\CI{20.26}{19.34}{21.74}
&
\CI{27.57}{26.54}{29.94}
&
\CI{71.79}{69.61}{73.89}
&
\CI{23.73}{22.73}{25.36}
&
\CI{30.22}{29.20}{32.86}
\\

ECG-CLIP~\cite{zhou2025diagnosis}
&
npj DM'25
&
\CI{65.22}{62.69}{67.68}
&
\CI{20.04}{19.01}{21.82}
&
\CI{28.35}{26.93}{30.42}
&
\CI{67.88}{65.41}{70.22}
&
\CI{22.85}{21.76}{24.21}
&
\CI{28.63}{27.51}{30.54}
&
\CI{73.92}{71.83}{75.72}
&
\CI{25.47}{24.34}{27.23}
&
\CI{\underline{31.72}}{30.54}{32.79}
\\

D-BETA~\cite{hung2025boosting}
&
ICML'25
&
\CI{66.43}{63.91}{68.81}
&
\CI{\underline{21.28}}{20.16}{22.73}
&
\CI{28.12}{26.74}{30.32}
&
\CI{69.46}{67.08}{71.68}
&
\CI{22.77}{21.72}{24.31}
&
\CI{\underline{29.42}}{28.23}{30.76}
&
\CI{73.68}{71.59}{75.52}
&
\CI{\underline{26.72}}{25.58}{27.96}
&
\CI{31.36}{30.25}{32.78}
\\

SGERA~\cite{chen2026sgera}
&
ICML'26
&
\CI{\underline{67.74}}{65.36}{69.12}
&
\CI{21.05}{19.98}{22.64}
&
\CI{\underline{29.21}}{27.85}{30.57}
&
\CI{\underline{70.33}}{68.05}{71.72}
&
\CI{22.58}{21.49}{24.16}
&
\CI{29.40}{28.26}{30.98}
&
\CI{\underline{74.37}}{72.42}{75.78}
&
\CI{26.41}{25.31}{27.82}
&
\CI{31.60}{30.53}{32.79}
\\

\midrule

\textbf{EchoBridge}
&
\textbf{Ours}
&
\CI{\textbf{69.40}}{67.03}{71.57}
&
\CI{\textbf{23.06}}{22.07}{24.48}
&
\CI{\textbf{30.78}}{29.82}{33.10}
&
\CI{\textbf{72.06}}{69.91}{74.15}
&
\CI{\textbf{24.76}}{23.73}{26.24}
&
\CI{\textbf{31.00}}{30.07}{33.00}
&
\CI{\textbf{75.99}}{74.02}{77.85}
&
\CI{\textbf{28.11}}{26.96}{29.89}
&
\CI{\textbf{32.98}}{32.25}{35.53}
\\

\bottomrule
\end{tabular}
\end{adjustbox}
\end{table*}

\begin{table*}[t]
\centering
\small
\setlength{\tabcolsep}{10pt}
\renewcommand{\arraystretch}{1.05}
\caption{
Source-only cross-center transfer from EchoNext-Mini to PKUPH and SHTMU. The ECG representation extractor and linear multi-label classifier are trained on EchoNext-Mini, and all model parameters and class-specific decision thresholds are fixed during target-cohort evaluation.
}
\label{tab:cross_domain_echonext_mini}

\begin{adjustbox}{max width=0.95\linewidth}
\begin{tabular}{@{}llccc|ccc@{}}
\toprule
\multirow{2}{*}{\textbf{Methods}}
& \multirow{2}{*}{\textbf{Ref.}}
& \multicolumn{3}{c}{\textbf{PKUPH}}
& \multicolumn{3}{c}{\textbf{SHTMU}} \\
\cmidrule(lr){3-5}
\cmidrule(lr){6-8}
&
& \textbf{AUROC}
& \textbf{AUPRC}
& \textbf{F1}
& \textbf{AUROC}
& \textbf{AUPRC}
& \textbf{F1} \\
\midrule

\multicolumn{8}{@{}l}{\textbf{ECG-only Self-Supervised Learning}} \\

SimCLR~\cite{chen2020simple}
& ICML'20
& \CI{70.78}{69.96}{71.61}
& \CI{13.46}{12.85}{14.19}
& \CI{22.26}{21.48}{23.31}
& \CI{70.28}{69.39}{71.16}
& \CI{17.33}{16.64}{18.18}
& \CI{24.21}{23.34}{25.38} \\

ST-MEM~\cite{na2024guiding}
& ICLR'24
& \CI{73.30}{72.51}{74.08}
& \CI{14.47}{13.84}{15.23}
& \CI{25.23}{24.42}{26.31}
& \CI{70.43}{69.55}{71.32}
& \CI{17.86}{17.18}{18.72}
& \CI{24.10}{23.24}{25.27} \\

HeartLang~\cite{jin2025reading}
& ICLR'25
& \CI{73.22}{72.43}{74.01}
& \CI{14.59}{13.95}{15.36}
& \CI{26.50}{25.67}{27.61}
& \CI{71.44}{70.57}{72.31}
& \CI{18.76}{18.05}{19.64}
& \CI{24.52}{23.63}{25.70} \\

\midrule
\multicolumn{8}{@{}l}{\textbf{ECG--Text Pretraining}} \\

CLIP~\cite{radford2021learning}
& ICML'21
& \CI{70.42}{69.57}{71.25}
& \CI{16.59}{15.92}{17.40}
& \CI{25.67}{24.84}{26.77}
& \CI{67.24}{66.31}{68.17}
& \CI{19.09}{18.35}{19.98}
& \CI{23.68}{22.78}{24.87} \\

SigLIP~\cite{zhai2023sigmoid}
& ICCV'23
& \CI{66.53}{65.64}{67.42}
& \CI{13.98}{13.34}{14.76}
& \CI{22.03}{21.23}{23.12}
& \CI{60.62}{59.61}{61.62}
& \CI{16.20}{15.50}{17.08}
& \CI{22.11}{21.20}{23.32} \\

PCME++~\cite{chun2023improved}
& ICLR'24
& \CI{66.92}{66.04}{67.81}
& \CI{14.59}{13.94}{15.38}
& \CI{23.58}{22.77}{24.68}
& \CI{64.51}{63.55}{65.47}
& \CI{15.73}{15.02}{16.61}
& \CI{21.69}{20.80}{22.88} \\

MERL-ECHO~\cite{wong2025contrastive}
& medRxiv'25
& \CI{76.30}{75.55}{77.05}
& \CI{16.54}{15.87}{17.36}
& \CI{26.40}{25.58}{27.50}
& \CI{72.47}{71.62}{73.32}
& \CI{18.75}{18.04}{19.62}
& \CI{\underline{25.69}}{24.80}{26.86} \\

ECG-CLIP~\cite{zhou2025diagnosis}
& npj DM'25
& \CI{74.09}{73.31}{74.87}
& \CI{14.60}{13.96}{15.38}
& \CI{24.39}{23.58}{25.48}
& \CI{72.41}{71.56}{73.27}
& \CI{19.20}{18.47}{20.10}
& \CI{24.74}{23.86}{25.91} \\

D-BETA~\cite{hung2025boosting}
& ICML'25
& \CI{\underline{76.78}}{76.04}{77.52}
& \CI{\underline{17.45}}{16.77}{18.27}
& \CI{\underline{27.13}}{26.31}{28.21}
& \CI{\underline{73.58}}{72.75}{74.41}
& \CI{\underline{20.00}}{19.26}{20.90}
& \CI{25.35}{24.46}{26.52} \\

SGERA~\cite{chen2026sgera}
& ICML'26
& \CI{72.36}{71.56}{73.16}
& \CI{15.94}{15.28}{16.74}
& \CI{25.82}{25.00}{26.91}
& \CI{69.83}{68.92}{70.74}
& \CI{18.31}{17.59}{19.19}
& \CI{24.69}{23.80}{25.87} \\

\midrule
\textbf{EchoBridge}
& \textbf{Ours}
& \CI{\textbf{77.92}}{77.20}{78.64}
& \CI{\textbf{18.98}}{18.28}{19.82}
& \CI{\textbf{27.39}}{26.57}{28.48}
& \CI{\textbf{74.11}}{73.29}{74.93}
& \CI{\textbf{22.08}}{21.31}{23.01}
& \CI{\textbf{28.16}}{27.24}{29.36} \\

\bottomrule
\end{tabular}
\end{adjustbox}
\end{table*}

\subsection{Target-Domain Cross-Center Frozen Linear Probing under Different Label Ratios}
Tables~\ref{tab:pkuph_ratio} and~\ref{tab:shtmu_ratio} evaluate cross-center transferability of ECG representations pretrained on EchoNext-Mini. For each method, the ECG representation extractor, including the encoder and output projection where applicable, is frozen, and a new linear multi-label classifier is trained independently on PKUPH or SHTMU using 1\%, 10\%, or 100\% of the target-cohort labels. Model selection and class-specific thresholds use only the corresponding validation split. This protocol measures linear accessibility under institutional and label-distribution shifts across target-domain supervision levels.

On PKUPH, EchoBridge achieves the highest point estimates for all metrics at every label ratio. Relative to the strongest baseline for each metric, it improves AUROC, AUPRC, and F1 by 1.77, 1.41, and 1.26 points with 1\% labels; 1.84, 1.28, and 1.42 points with 10\%; and 1.46, 1.32, and 1.34 points with 100\%, respectively. The gains at 1\% and 10\% indicate adaptability under limited target-center supervision. On SHTMU, EchoBridge also achieves the highest point estimates across all metric--budget combinations. Its AUROC, AUPRC, and F1 gains are 1.66, 1.78, and 1.57 points with 1\% labels; 1.73, 1.69, and 1.58 points with 10\%; and 1.62, 1.39, and 1.26 points with 100\%, respectively. Consistent gains across both cohorts support target-domain adaptability and label efficiency under
institutional and prevalence shifts.

\subsection{Source-Only Cross-Center Transfer}
Table~\ref{tab:cross_domain_echonext_mini} evaluates source-only cross-center transfer by training the linear classifier and selecting class-specific thresholds exclusively on EchoNext-Mini, then applying the fixed representation extractor, classifier, and thresholds to findings shared with each target cohort. EchoBridge exceeds the strongest competing method for each metric by 1.14 AUROC, 1.53 AUPRC, and 0.26 F1 points on PKUPH, and by 0.53, 2.08, and 2.47 points on SHTMU, respectively. On PKUPH, the larger AUROC and AUPRC gains relative to F1 suggest improved score ranking with limited change under the transferred source-domain thresholds. On SHTMU, the larger AUPRC and F1 gains indicate improved positive-finding discrimination under the shifted prevalence distribution. These results complement target-domain linear probing and demonstrate direct cross-center transfer without target-domain training, calibration, or threshold adjustment.

\begin{table*}[t]
\centering
\scriptsize
\setlength{\tabcolsep}{4pt}
\renewcommand{\arraystretch}{1.05}

\caption{
Matched label-only control and component-wise ablation of CSPP,
FAAM, and SRR on EchoNext-Mini. Performance is evaluated using
prompt-based classifier-free inference and frozen linear probing.
}
\label{tab:ablation_variants}

\begin{adjustbox}{max width=\textwidth}
\begin{tabular}{@{}lcccc|ccc|ccc|ccc|ccc@{}}
\toprule
\multirow{2}{*}{\textbf{Configuration}}
& \multirow{2}{*}{
    \shortstack{
    \textbf{ECG--Text Align.}\\
    \textbf{+ Proto.}}}
& \multirow{2}{*}{\textbf{CSPP}}
& \multicolumn{2}{c}{\textbf{APBC}}
& \multicolumn{3}{c}{\textbf{Prompt-based}}
& \multicolumn{3}{c}{\textbf{1\% Linear Probing}}
& \multicolumn{3}{c}{\textbf{10\% Linear Probing}}
& \multicolumn{3}{c}{\textbf{100\% Linear Probing}} \\
\cmidrule(lr){4-5}
\cmidrule(lr){6-8}
\cmidrule(lr){9-11}
\cmidrule(lr){12-14}
\cmidrule(lr){15-17}
&
&
& \textbf{FAAM}
& \textbf{SRR}
& \textbf{AUROC}
& \textbf{AUPRC}
& \textbf{F1}
& \textbf{AUROC}
& \textbf{AUPRC}
& \textbf{F1}
& \textbf{AUROC}
& \textbf{AUPRC}
& \textbf{F1}
& \textbf{AUROC}
& \textbf{AUPRC}
& \textbf{F1} \\
\midrule

\textbf{Label-only BCE}
& $\times$
& $\times$
& $\times$
& $\times$
& \multicolumn{3}{c|}{--}
& 64.79
& 18.06
& 24.53
& 73.79
& 26.16
& 31.07
& 76.87
& 29.89
& 33.96 \\

\midrule

\textbf{Align. + Proto.}
& $\checkmark$
& $\times$
& $\times$
& $\times$
& 67.18
& 18.83
& 24.09
& 68.48
& 21.90
& 28.00
& 74.11
& 27.20
& 31.80
& 76.61
& 30.18
& 34.08 \\

\textbf{+ CSPP}
& $\checkmark$
& $\checkmark$
& $\times$
& $\times$
& 71.05
& 22.39
& 27.74
& 70.08
& 23.58
& 29.27
& 75.16
& 28.36
& 32.70
& 77.42
& 31.06
& 34.76 \\

\textbf{+ APBC}
& $\checkmark$
& $\times$
& $\checkmark$
& $\checkmark$
& 73.36
& 24.71
& 30.02
& 70.88
& 24.51
& 29.94
& 75.69
& 29.03
& 33.18
& 77.83
& 31.58
& 35.13 \\

\textbf{+ CSPP + FAAM}
& $\checkmark$
& $\checkmark$
& $\checkmark$
& $\times$
& 74.02
& 25.18
& 30.46
& 71.69
& 25.40
& 30.52
& 76.23
& 29.66
& 33.58
& 78.24
& 32.07
& 35.42 \\

\textbf{+ CSPP + SRR}
& $\checkmark$
& $\checkmark$
& $\times$
& $\checkmark$
& 73.91
& 25.04
& 30.31
& 71.43
& 25.12
& 30.28
& 76.06
& 29.47
& 33.40
& 78.11
& 31.92
& 35.27 \\

\midrule

\textbf{EchoBridge}
& $\checkmark$
& $\checkmark$
& $\checkmark$
& $\checkmark$
& \textbf{75.73}
& \textbf{26.79}
& \textbf{31.83}
& \textbf{72.77}
& \textbf{26.53}
& \textbf{31.32}
& \textbf{76.94}
& \textbf{30.45}
& \textbf{34.13}
& \textbf{78.79}
& \textbf{32.66}
& \textbf{35.82} \\

\bottomrule
\end{tabular}
\end{adjustbox}
\end{table*}

\subsection{Ablation Study}

Table~\ref{tab:ablation_variants} reports a matched label-only control and ablations of CSPP and the APBC refinements FAAM and SRR. All ECG--text variants combine bidirectional alignment with class-prototype supervision, using Align.+Proto. as the reference. The label-only BCE control tests whether structured-label supervision alone explains the gains. It uses the same ECG encoder, 256-dimensional projection, data split, optimizer, batch size, epoch budget, and schedule as the ECG--text variants, with pretraining driven solely by structured multi-label BCE. Its encoder and projection are then frozen, and a new linear classifier is trained with 1\%, 10\%, or 100\% of the probe labels. Prompt-based inference is reported only for ECG--text variants because it requires a text-aligned space.

Align.+Proto. outperforms the label-only control under 1\% and 10\% probing, indicating greater linear accessibility with limited supervision, while their 100\% results are comparable. CSPP improves prompt-based inference and frozen probing across all budgets, and complete APBC without CSPP improves every metric. With CSPP, both FAAM and SRR outperform CSPP alone, with slightly larger gains from FAAM. EchoBridge performs best across all protocols, supporting their complementary contributions beyond Align.+Proto. and matched label supervision.

\subsection{Finding-Specific Performance across Prevalence Levels}
\begin{figure}[t]
\centering
\includegraphics[width=\linewidth]{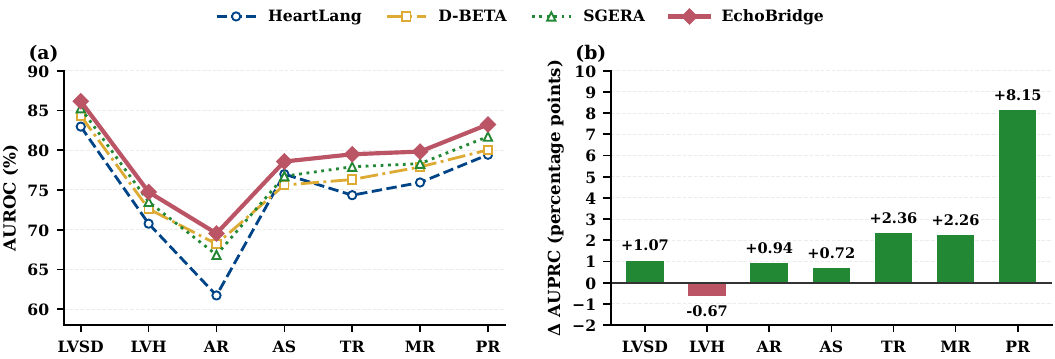}
\caption{
Finding-specific performance on EchoNext-Mini under 100\% frozen linear probing. (a) AUROC by method for each echocardiography-derived finding. (b) EchoBridge's AUPRC difference from the strongest baseline for each finding in percentage points; positive values indicate gains.
}

\label{Figure5}
\end{figure}

As shown in Figure~\ref{Figure5}, under 100\% frozen linear probing on EchoNext-Mini, EchoBridge achieves the highest AUROC point estimate for all seven findings, with the largest gains for moderate-or-severe aortic stenosis and tricuspid regurgitation at 1.59 and 1.57 points, respectively. It also obtains the highest AUPRC for six findings, improving pulmonic and tricuspid regurgitation by 8.15 and 2.36 points, respectively, while its LVH AUPRC is 0.67 points lower.

Appendix~\ref{additional_results} reports finding-specific AUROC and AUPRC on EchoNext-Mini, PKUPH, and SHTMU under the corresponding 100\% frozen linear-probing protocols. On both target cohorts, EchoBridge achieves the highest point estimates for every finding across ventricular function, chamber abnormalities, and valvular disease. These cross-institutional gains include several low-prevalence valvular abnormalities. Their varying magnitudes indicate condition-specific benefits, while estimates based on few positive test cases require cautious interpretation under severe class imbalance.

\section{Limitations and Ethical Considerations}
The retrospective study covers a limited number of institutions; prospective multicenter validation across populations, acquisition systems, and clinical workflows remains necessary, particularly for low-prevalence valvular abnormalities. Performance may also vary across demographic groups, ECG devices, acquisition protocols, and care settings. All private data were de-identified and processed in access-controlled institutional environments. The PKUPH and SHTMU cohorts were retrospectively collected under approvals from the Institutional Review Boards of Peking University People's Hospital (Approval No.~2024PHB428-001) and the Second Hospital of Tianjin Medical University (Approval No.~KY2025K386), respectively. EchoBridge is intended for research and clinician-assisted screening, with predictions interpreted by qualified professionals alongside other clinical evidence. False-negative predictions, particularly for rare abnormalities, may delay further assessment. Clinical deployment requires site-specific validation, calibration, monitoring, regulatory review, and explicit human oversight.

\section{Conclusion}
We presented EchoBridge, an ECG--echocardiography text alignment framework for transferable representations of echocardiography-derived cardiac findings. It combines Complementary Shared--Private Projection and Adaptive Prototype Boundary Calibration to address cross-modal interference and long-tailed distributions. Across EchoNext-Mini and two independent cohorts, EchoBridge outperforms representative ECG-only and ECG--text baselines across classifier-free inference, in-domain and target-domain frozen probing, source-only transfer, and finding-specific evaluation.

\section*{Generative AI Usage}
LLMs assisted with auxiliary coding, translation, editing, and drafting definition-only prompts for echocardiography-derived findings. GPT-5.5 Thinking generated initial candidates, which senior cardiologists reviewed and standardized against predefined label definitions before testing. The authors independently developed the methodology, conducted experiments, interpreted results, and assume full responsibility for the manuscript.

\bibliographystyle{ACM-Reference-Format}
\bibliography{sample-sigconf}


\appendix
\section{Acknowledgments}
As an informal quality check, five senior cardiologists inspected a randomly sampled subset of mappings from structured cardiac findings to standardized conclusion-style summaries. Each sampled summary was presented with its corresponding finding vector. The reviewers provided qualitative feedback that the sampled summaries were broadly clinically plausible and semantically consistent with the encoded findings. This review was not designed as a formal annotation, quantitative validation, or inter-rater agreement study.

The expert panel comprised Qinghao Zhao (Peking University People's Hospital, Beijing, China), Guanyu Mu (The Second Hospital of Tianjin Medical University, Tianjin, China), Xingliang Wu (Tianjin Institute of Cardiology, Tianjin, China), Xinxin Di (The First Affiliated Hospital of USTC, Hefei, China), and Jing Zhao (The First Affiliated Hospital of Anhui Medical University, Hefei, China).

\section{Related Work}
\subsection{Direct Echocardiography Supervision}
Echocardiography-derived supervision is widely used for ECG-based structural heart disease assessment. Early studies paired ECG and echocardiography data to detect ventricular systolic dysfunction or reduced ejection fraction~\cite{attia2019screening,yao2021artificial}, followed by work on specific abnormalities such as aortic stenosis and left-sided valvular disease~\cite{kwon2020deep,elias2022deep}. These studies established supervised ECG-to-echo prediction using echocardiographic measurements or diagnoses as screening targets. More recent work expanded to broader structural heart disease assessment. rECHOmmend~\cite{ulloa2022rechommend} used a composite endpoint to identify patients at risk of undiagnosed echocardiography-detectable disease, while Fujiki et al.~\cite{fujiki2025deep} studied multi-label prediction across ventricular function, chamber structure, and valvular abnormalities. EchoNext scaled this direction with paired examinations and public resources for evaluating echo-confirmed structural heart disease detection~\cite{poterucha2025detecting,hughes2026echonext}. Related studies also examined ECG images and single-lead recordings~\cite{dhingra2025ensemble,aminorroaya2025development}. These methods use predefined echocardiography-derived targets, motivating compositional ECG--echocardiography alignment organized around predefined findings and their co-occurrence patterns.

\subsection{ECG--Report Representation Learning}

ECG--text pretraining is a major approach to language-supervised ECG representation learning. METS~\cite{li2024frozen} and ETP~\cite{liu2024etp} learn shared embeddings from paired ECGs and machine-generated or clinical reports for zero-shot classification and label-efficient evaluation. MERL~\cite{liu2024zero} uses clinical knowledge-enhanced prompts at inference, while ECG-CLIP~\cite{zhou2025diagnosis} scales CLIP-style supervision to large ECG--report datasets for zero-shot diagnosis across cardiac conditions. Multimodal ECG pretraining further combines reports, EHR records, PPG, and cardiac imaging~\cite{lalam2023ecg,fang2025ppgflowecg,nie2025anyppg,fang2026ecgflowcmr}. Recent methods introduce richer interaction and structured supervision~\cite{jin2026ecg}. DERI~\cite{chen2025deri} combines multiple alignment objectives with mutual feature reconstruction, ESI~\cite{yu2024ecg} enriches reports with LLM-generated cardiological descriptions, and K-MERL~\cite{liu2025knowledge} extracts structured knowledge from free-text reports while supporting arbitrary-lead inputs. D-BETA~\cite{hung2025boosting} integrates masked ECG--text autoencoding with discriminative contrastive learning, while SGERA~\cite{chen2026sgera} uses Stein-guided alignment to address structural and statistical modality discrepancies. These studies establish ECG--report alignment for transferable ECG representation learning. Their supervision primarily comes from ECG reports, diagnostic statements, and broader clinical records, emphasizing rhythm, conduction, waveform morphology, and general ECG diagnoses.

\subsection{ECG--Echocardiography Text Alignment}
Recent studies use paired ECGs and echocardiography reports as cross-modal supervision for structural cardiac representation learning. MERL-ECHO applies CLIP-style contrastive pretraining to encode 12-lead ECGs and echocardiography reports in a shared space for zero-shot structural heart disease prediction~\cite{wong2025contrastive}. Wearable-Echo-FM extends this paradigm to single-lead ECGs and evaluates label-efficient fine-tuning for left ventricular systolic dysfunction, diastolic dysfunction, and composite structural heart disease~\cite{knight2026wearable}. These studies establish ECG--echocardiography text alignment as a viable approach for transferring structural cardiac information to ECG encoders and improving label efficiency. However, global cross-modal alignment may entangle modality-specific factors, while imbalanced finding distributions provide uneven class-level supervision. EchoBridge addresses these limitations through shared--private projection and class-aware geometric organization of the normalized representation space.

\begin{algorithm*}[t]
\caption{End-to-End Training of EchoBridge}
\label{alg:echobridge}
\small
\algrenewcommand\algorithmicindent{0.8em}
\begin{algorithmic}[1]
\Require Labeled paired data
$\mathcal{D}=\{(x_i,t_i,\mathbf{y}_i)\}$;
training epochs $E$;
number of cardiac findings $C$;
training-set positive rates
$\mathbf{r}=\{r_c\}_{c=1}^{C}$;
alignment temperature $\tau$;
margin parameters $(m_0,m_{\min},m_{\max})$;
numerical constant $\epsilon$;
Riesz exponent $q$;
repulsion weight $\lambda_r$.
\Ensure Optimized model parameters $\Theta$.

\State Initialize the learnable prototype matrix
$P=[p_1,\ldots,p_C]^{\top}\in\mathbb{R}^{C\times d}$
and learnable logit-scale parameter $s$.
\State Compute the normalization factor
$\kappa\gets\frac{1}{C}\sum_{c=1}^{C}
\frac{\operatorname{median}(\mathbf{r})}{r_c+\epsilon}$.
\State Compute the fixed class-specific margins
$m_c\gets\operatorname{clip}\!\left(
m_0\frac{\operatorname{median}(\mathbf{r})}{r_c+\epsilon}
\frac{1}{\kappa},
m_{\min},m_{\max}\right)$
for $c=1,\ldots,C$.

\For{$e=1,\ldots,E$}
    \For{each mini-batch
    $\{(x_i,t_i,\mathbf{y}_i)\}_{i=1}^{B}\sim\mathcal{D}$}

        \State Encode the paired inputs:
        $h_{e,i}\gets f_e(x_i)$ and
        $h_{t,i}\gets f_t(t_i)$.

        \State Obtain the shared and private ECG projections:
        $z_{e,i}^{s}\gets\phi_e^{s}(h_{e,i})$ and
        $z_{e,i}^{p}\gets\phi_e^{p}(h_{e,i})$.

        \State Obtain the shared and private text projections:
        $z_{t,i}^{s}\gets\phi_t^{s}(h_{t,i})$ and
        $z_{t,i}^{p}\gets\phi_t^{p}(h_{t,i})$.

        \State Construct normalized shared and private copies and compute
        the within-modality orthogonality loss
        $\mathcal{L}_{\mathrm{orth}}$ using Eq.~(4).

        \State Normalize the shared representations and class prototypes:
        $\hat{z}_{e,i}^{s}\gets
        z_{e,i}^{s}/\lVert z_{e,i}^{s}\rVert_2$,
        $\hat{z}_{t,i}^{s}\gets
        z_{t,i}^{s}/\lVert z_{t,i}^{s}\rVert_2$, and
        $\hat{p}_{c}\gets p_c/\lVert p_c\rVert_2$.

        \State Compute the cross-modal similarity matrix and symmetric
        alignment loss $\mathcal{L}_{\mathrm{align}}$
        using Eqs.~(6)--(7).

        \State Compute ECG--prototype and text--prototype cosine similarities
        $u_{i,c}^{e}=\langle\hat{z}_{e,i}^{s},\hat{p}_c\rangle$
        and
        $u_{i,c}^{t}=\langle\hat{z}_{t,i}^{s},\hat{p}_c\rangle$
        using Eq.~(10).

        \State Set $\gamma\gets\exp(s)$ and compute the margin-adjusted logits
        $\tilde{\ell}_{i,c}^{e}$ and $\tilde{\ell}_{i,c}^{t}$
        by applying $m_c$ only to positive sample--prototype pairs
        according to Eq.~(14).

        \State Compute the multi-label prototype loss
        $\mathcal{L}_{\mathrm{proto}}
        \gets
        \operatorname{BCE}(\tilde{\ell}^{e},\mathbf{y})
        +
        \operatorname{BCE}(\tilde{\ell}^{t},\mathbf{y})$
        using Eq.~(15).

        \State Compute the spherical Riesz repulsion loss
        $\mathcal{L}_{\mathrm{riesz}}$
        over all normalized class-prototype pairs using Eq.~(16).

        \State Compute
        $\mathcal{L}_{\mathrm{apbc}}
        \gets
        \mathcal{L}_{\mathrm{proto}}
        +\lambda_r\mathcal{L}_{\mathrm{riesz}}$
        and
        $\mathcal{L}
        \gets
        \mathcal{L}_{\mathrm{align}}
        +\mathcal{L}_{\mathrm{orth}}
        +\mathcal{L}_{\mathrm{apbc}}$.

        \State Update
        $f_e$, $f_t$,
        $\phi_e^{s}$, $\phi_e^{p}$,
        $\phi_t^{s}$, $\phi_t^{p}$,
        $P$, and $s$
        using AdamW and $\nabla_{\Theta}\mathcal{L}$.

    \EndFor
\EndFor

\State \Return ECG encoder $f_e$ and shared ECG projection head $\phi_e^{s}$.
\end{algorithmic}
\end{algorithm*}

\begin{table*}[t]
\centering
\caption{
Layer-wise configuration of the one-dimensional ResNet-18 ECG encoder for an input ECG of shape $12\times5000$. Each residual stage contains two basic blocks, and the reported stride applies to the first block of each stage.
}
\label{tab:ecg_architecture}

\small
\setlength{\tabcolsep}{8pt}
\renewcommand{\arraystretch}{1.05}

\begin{adjustbox}{max width=\textwidth}
\begin{tabular}{@{}lcccccc@{}}
\toprule
\textbf{Component}
& \textbf{Blocks}
& \textbf{Input channels}
& \textbf{Output channels}
& \textbf{Kernel size}
& \textbf{First-block stride}
& \textbf{Temporal length} \\
\midrule

Stem convolution
& 1
& 12
& 64
& 7
& 2
& 2500 \\

Residual stage 1
& 2
& 64
& 64
& 3
& 1
& 2500 \\

Residual stage 2
& 2
& 64
& 128
& 3
& 2
& 1250 \\

Residual stage 3
& 2
& 128
& 256
& 3
& 2
& 625 \\

Residual stage 4
& 2
& 256
& 512
& 3
& 2
& 313 \\

Adaptive average pooling
& 1
& 512
& 512
& --
& --
& 1 \\

\bottomrule
\end{tabular}
\end{adjustbox}
\end{table*}

\section{Implementation Details of EchoBridge}
\subsection{Algorithm of EchoBridge}
Algorithm~\ref{alg:echobridge} summarizes EchoBridge's end-to-end training, including shared--private projection, within-modality orthogonality, bidirectional ECG--text alignment, frequency-adaptive prototype calibration, spherical Riesz repulsion, and joint optimization.

\subsection{Encoder and Projection Architectures}

\paragraph{\bfseries ECG Encoder.}
We use a one-dimensional ResNet-18 initialized from scratch to encode each 10-second, 12-lead ECG recording with input shape $12\times5000$. The encoder follows the four-stage ResNet-18 architecture, with all two-dimensional operations replaced by one-dimensional counterparts. We remove the max-pooling layer after the stem convolution to preserve temporal resolution. The stem comprises a bias-free $\operatorname{Conv1D}(12,64,7,2,3)$ layer followed by batch normalization and ReLU.

Each basic residual block contains two 3-tap convolutions:
\begin{equation}
\operatorname{Conv1D}(k=3)
\rightarrow
\operatorname{BN}
\rightarrow
\operatorname{ReLU}
\rightarrow
\operatorname{Conv1D}(k=3)
\rightarrow
\operatorname{BN}.
\end{equation}
The first block of each stage except the first uses stride 2 to reduce temporal resolution and increase channels. Dimension-changing shortcuts use a bias-free $1\times1$ convolution followed by batch normalization. A ReLU is applied after adding the residual and shortcut paths. Adaptive average pooling produces the 512-dimensional ECG representation $h_e$. Table~\ref{tab:ecg_architecture} summarizes the architecture.

\paragraph{\bfseries Text Encoder.}
We use the pretrained MedCPT-Article-Encoder~\cite{jin2023medcpt}, a BERT-base model with 12 Transformer layers, a hidden size of 768, 12 self-attention heads, and a 3,072-dimensional feed-forward layer. It uses GELU, 0.1 dropout, a 30,522-token vocabulary, and a maximum sequence length of 512. The 768-dimensional \texttt{pooler\_output} serves as the global text representation $h_t$. All parameters are jointly fine-tuned during cross-modal pretraining.

\paragraph{\bfseries Shared and Private Projections.}
Four independent two-layer heads map the ECG and text representations into 256-dimensional shared and auxiliary private projections:
\begin{align}
\phi_e^{s},\, \phi_e^{p}
&:
\operatorname{Linear}(512,256)
\rightarrow \operatorname{GELU}
\rightarrow \operatorname{Linear}(256,256),\\
\phi_t^{s},\, \phi_t^{p}
&:
\operatorname{Linear}(768,256)
\rightarrow \operatorname{GELU}
\rightarrow \operatorname{Linear}(256,256).
\end{align}
Shared projections are $\ell_2$-normalized before cross-modal alignment and prototype supervision. Private projections remain unnormalized, while normalized copies of both branches compute the within-modality orthogonality loss.

\section{Additional Dataset Details}
EchoNext-Mini labels follow the published structured definitions of the original dataset. For PKUPH and SHTMU, echocardiography-derived findings were extracted from physician-authored reports using regular-expression-based natural language processing pipelines. The pipelines normalized synonymous clinical terms, handled negation and uncertainty expressions, parsed severity modifiers, and mapped report statements to predefined binary findings. Senior cardiologist Qinghao Zhao subsequently reviewed all extracted labels from both cohorts for clinical consistency.

\subsection{EchoNext-Mini Dataset}
EchoNext-Mini~\cite{hughes2026echonext} is a de-identified public subset of EchoNext derived from routine clinical data at Columbia University Irving Medical Center, including Columbia and Allen hospitals. It contains 100,000 10-second, 12-lead ECGs from 36,286 adults aged at least 18 years. Recordings were acquired between 2008 and 2022 at 250~Hz and linked to transthoracic echocardiograms performed within one year. Each record includes demographic, acquisition, and automated ECG measurement metadata; ages above 90 were capped at 90 for de-identification.

Echocardiography-derived labels were constructed from structured report fields and quantitative measurements, including left ventricular ejection fraction (LVEF), ventricular wall thickness, pulmonary artery systolic pressure, tricuspid regurgitation velocity, valvular severity, right ventricular systolic function, and pericardial effusion. The dataset provides 11 component abnormalities and a composite moderate-or-greater structural heart disease label. Binary criteria include LVEF $\leq 45\%$ for left ventricular systolic dysfunction, maximum septal or posterior wall thickness $\geq 13$~mm for increased wall thickness, and moderate-or-severe grading for aortic stenosis and aortic, mitral, tricuspid, and pulmonic regurgitation. Positive labels require the ECG to precede the echocardiogram by at most one year. Echocardiograms with prosthetic valves, missing LVEF, or unavailable wall-thickness measurements were excluded.

We evaluate seven findings: left ventricular systolic dysfunction, left ventricular hypertrophy, and moderate-or-severe aortic regurgitation, aortic stenosis, tricuspid regurgitation, mitral regurgitation, and pulmonic regurgitation. Prevalence is strongly imbalanced: left ventricular systolic dysfunction and hypertrophy each occur in approximately 24\% of samples, whereas pulmonic and aortic regurgitation occur in approximately 0.8\%--1.3\%.

\subsection{PKUPH Dataset}
The PKUPH cohort was retrospectively assembled from longitudinal ECG and transthoracic echocardiography data collected at Peking University People's Hospital from June 2015 to May 2023. Of 74,220 screened individuals, the final cohort included 20,768 patients and 27,158 ECG recordings. ECG--echocardiography pairs were matched by prioritizing same-day examinations; otherwise, the temporally closest ECG within a $\pm10$-day window was selected. Pairs outside this window and ECGs with corrupted waveforms, incomplete leads, or inconsistent metadata were excluded.

The mean age was $61.6 \pm 14.4$ years, and 9,193 patients (44.3\%) were male. Cardiac findings were extracted from physician-authored echocardiography reports using rule-based natural language processing. The cohort includes broader diastolic-function- and wall-motion-related findings than EchoNext-Mini, including left ventricular diastolic dysfunction and wall-motion abnormality, and exhibits a distinct prevalence distribution, enabling cross-center evaluation under institutional and label-prevalence shifts.

\subsection{SHTMU Dataset}
The SHTMU cohort was retrospectively assembled from longitudinal ECG and transthoracic echocardiography data collected at the Second Hospital of Tianjin Medical University between January and December 2024. Among 479,089 individuals screened from the institutional repository, 462,468 lacked available echocardiography data. The resulting cohort included 16,621 patients and 18,588 ECG recordings. ECG--echocardiography pairs were constructed using the same temporal matching strategy as for PKUPH. Same-day examinations were prioritized; when no same-day ECG was available, the temporally closest ECG within $\pm10$ days of the echocardiography report was selected. Pairs outside this window were excluded, together with ECGs containing corrupted waveforms, incomplete lead data, or inconsistent metadata.

The patients had a mean age of $65.3 \pm 13.7$ years, and 9,654 were male. Compared with PKUPH, SHTMU showed a substantially different finding-prevalence profile, characterized by a high prevalence of left atrial enlargement and a low prevalence of findings such as right atrial enlargement. The cohort therefore supports geographically independent cross-center evaluation under concurrent institutional and label-prevalence shifts.

\begin{table*}[t]
\centering
\caption{
Definition-only prompts used to construct textual prototypes for prompt-based classifier-free inference on EchoNext-Mini. Each prompt describes the corresponding echocardiography-derived finding through target-defining anatomical, functional, hemodynamic, severity-related, or quantitative characteristics.
}
\label{tab:prompts}
\small
\setlength{\tabcolsep}{5pt}
\renewcommand{\arraystretch}{1.15}
\begin{tabularx}{\textwidth}{
    >{\raggedright\arraybackslash}p{0.25\textwidth}
    >{\raggedright\arraybackslash}X}
\toprule
\textbf{Cardiac finding} & \textbf{Text prompt} \\
\midrule

Left ventricular systolic dysfunction
&
The echocardiographic examination demonstrates impaired global left
ventricular contractile function, typically characterized by a left
ventricular ejection fraction of 45\% or lower.
\\

Left ventricular hypertrophy
&
The echocardiographic examination demonstrates increased left ventricular
myocardial wall thickness or mass, typically involving thickening of the
interventricular septum, the left ventricular posterior wall, or both.
\\

Moderate or severe aortic regurgitation
&
The echocardiographic examination demonstrates moderate or severe diastolic
regurgitant blood flow across the aortic valve from the aorta into the left
ventricle, reflecting clinically significant aortic valve incompetence.
\\

Moderate or severe aortic stenosis
&
The echocardiographic examination demonstrates moderate or severe narrowing
and restricted opening of the aortic valve, resulting in hemodynamically
significant obstruction of systolic blood flow from the left ventricle into
the aorta, typically accompanied by increased transvalvular velocity or
pressure gradient and reduced valve area.
\\

Moderate or severe tricuspid regurgitation
&
The echocardiographic examination demonstrates moderate or severe systolic
regurgitant blood flow across the tricuspid valve from the right ventricle
into the right atrium, reflecting clinically significant tricuspid valve
incompetence.
\\

Moderate or severe mitral regurgitation
&
The echocardiographic examination demonstrates moderate or severe systolic
regurgitant blood flow across the mitral valve from the left ventricle into
the left atrium, reflecting clinically significant mitral valve
incompetence.
\\

Moderate or severe pulmonic regurgitation
&
The echocardiographic examination demonstrates moderate or severe diastolic
regurgitant blood flow across the pulmonic valve from the pulmonary artery
into the right ventricle, reflecting clinically significant pulmonic valve
incompetence.
\\

\bottomrule
\end{tabularx}
\end{table*}

\section{Additional Experimental Details}

\subsection{Detailed Evaluation Protocols}
\label{app:evaluation_protocols}

\paragraph{\bfseries Prompt-Based Classifier-Free Inference.}
We freeze the pretrained ECG encoder, shared ECG projection head, text encoder, and shared text projection head, and perform inference without training a downstream classifier. Each cardiac finding is represented by a fixed definition-only prompt, which the text branch encodes into an $\ell_2$-normalized textual prototype. Each ECG is mapped into the normalized shared space, and its cosine similarity with each textual prototype serves as the corresponding class-wise prediction score. Class-specific thresholds are selected on the EchoNext-Mini validation set and fixed before test evaluation. Because all evaluated findings are incorporated during pretraining through standardized summaries and class-prototype supervision, this protocol evaluates the semantic accessibility of pretraining-seen findings. Generalization to unseen disease categories lies outside its scope.

\paragraph{\bfseries In-Domain Frozen Linear Probing.}
We freeze the pretrained ECG encoder and output projection and train a linear multi-label classifier on EchoNext-Mini using 1\%, 10\%, or 100\% of the available training labels. This protocol evaluates the linear accessibility of the frozen representations across different label budgets. Label subsets are sampled using fixed random seeds shared across methods. Model selection and class-specific threshold determination use only the EchoNext-Mini validation split, and performance is reported on the independent patient-disjoint test set. All methods use identical label subsets, classifier architectures, optimization settings, and evaluation procedures.

\paragraph{\bfseries Target-Domain Cross-Center Frozen Linear Probing.}
We evaluate the adaptability of EchoNext-Mini-pretrained representations to the independent PKUPH and SHTMU cohorts. For each method, the pretrained ECG representation extractor, including the ECG encoder and output projection where applicable, is frozen. A new linear multi-label classifier is trained independently on each target cohort using 1\%, 10\%, or 100\% of its available training labels. Label subsets are sampled using fixed random seeds shared across methods. Model selection and class-specific threshold determination use only the validation split of the corresponding target cohort, and performance is reported on its patient-disjoint test set. This protocol evaluates the linear accessibility of the frozen representations under institutional and label-distribution shifts across different levels of target-domain supervision.

\paragraph{\bfseries Source-Only Cross-Center Transfer.}
We evaluate direct cross-center transfer without target-domain training or calibration. A linear multi-label classifier is trained on frozen ECG representations using the complete EchoNext-Mini training set, while model selection and class-specific threshold determination use only the EchoNext-Mini validation split. The ECG encoder, output projection, linear classifier, and decision thresholds are then fixed and applied unchanged to PKUPH and SHTMU. Evaluation is restricted to cardiac findings shared between EchoNext-Mini and each target cohort. Target-domain samples are excluded from representation learning, classifier training, model selection, calibration, and threshold determination.

\begin{table}[t]
\centering
\caption{Hyperparameters used for EchoBridge pretraining.}
\label{tab:alignment_hyperparameters}
\small
\setlength{\tabcolsep}{3pt}
\renewcommand{\arraystretch}{1.08}
\begin{tabularx}{\columnwidth}{@{}Xr@{}}
\toprule
\textbf{Hyperparameter} & \textbf{Value} \\
\midrule

\multicolumn{2}{@{}l}{\textit{\textbf{Frequency-adaptive angular margin}}} \\
Base margin $m_0$ & $0.20$ rad \\
Minimum margin $m_{\min}$ & $0.05$ rad \\
Maximum margin $m_{\max}$ & $0.50$ rad \\

\multicolumn{2}{@{}l}{\textit{\textbf{Spherical Riesz regularization}}} \\
Riesz exponent $q$ & $2.0$ \\
Regularization weight $\lambda_r$ & $0.05$ \\

\multicolumn{2}{@{}l}{\textit{\textbf{Training configuration}}} \\
Learning rate & $1\times 10^{-5}$ \\
Batch size & $64$ \\
Max epochs & $15$ \\
Random seed & $42$ \\
Data-loader workers & $8$ \\
Checkpoint interval & $5$ epochs \\

\bottomrule
\end{tabularx}
\end{table}

\subsection{Pretraining Hyperparameters}
Table~\ref{tab:alignment_hyperparameters} summarizes the optimization and alignment hyperparameters used for EchoBridge pretraining.

\begin{table*}[t]
\centering
\small
\setlength{\tabcolsep}{5pt}
\renewcommand{\arraystretch}{1.05}
\caption{
Finding-specific AUROC and AUPRC on EchoNext-Mini under 100\% frozen linear probing. P/N denotes the numbers of positive and negative test samples for each echocardiography-derived finding. Mod./Sev. denotes moderate-or-severe disease.
}
\label{tab:disease_specific_auroc}
\begin{adjustbox}{max width=\textwidth,center}
\begin{tabular}{@{}lcccccccc@{}}
\toprule
\multicolumn{1}{c}{\multirow{2}{*}{\textbf{Methods}}} 
& \multirow{2}{*}{\textbf{Ref.}}
& \textbf{LVSD}
& \textbf{LVH}
& \textbf{Mod./Sev. AR}
& \textbf{Mod./Sev. AS}
& \textbf{Mod./Sev. TR}
& \textbf{Mod./Sev. MR}
& \textbf{Mod./Sev. PR} \\
& 
& {\scriptsize P/N=4833/15167}
& {\scriptsize P/N=4954/15046}
& {\scriptsize P/N=268/19732}
& {\scriptsize P/N=826/19174}
& {\scriptsize P/N=2172/17828}
& {\scriptsize P/N=1715/18285}
& {\scriptsize P/N=154/19846} \\
\midrule

\multicolumn{9}{@{}l}{\textbf{ECG-only Self-Supervised Learning}} \\
SimCLR~\cite{chen2020simple} 
& ICML'20
& 75.18/50.06 & 65.64/35.39 & 62.00/2.19 & 65.60/7.87 & 67.95/20.33 & 69.65/17.91 & 74.87/5.99 \\

ST-MEM~\cite{na2024guiding} 
& ICLR'24
& 78.73/54.72 & 68.53/39.32 & 59.17/1.91 & 74.00/11.96 & 70.77/22.42 & 72.41/18.96 & 74.54/2.73 \\

HeartLang~\cite{jin2025reading} 
& ICLR'25
& 82.97/63.16 & 70.76/41.92 & 61.73/2.31 & \underline{76.99}/\underline{14.33} & 74.34/28.05 & 75.94/22.36 & 79.43/5.44 \\

\midrule
\multicolumn{9}{@{}l}{\textbf{ECG-Text Pretraining}} \\
CLIP~\cite{radford2021learning} 
& ICML'21
& 79.20/57.87 & 67.01/38.22 & 63.34/2.19 & 67.19/8.42 & 69.65/22.41 & 72.64/20.83 & 74.23/5.04 \\

SigLIP~\cite{zhai2023sigmoid} 
& ICCV'23
& 78.86/57.46 & 67.14/38.24 & 64.18/2.13 & 67.08/8.30 & 70.40/22.64 & 71.89/19.89 & 77.25/4.41 \\

PCME++~\cite{chun2023improved} 
& ICLR'24
& 76.63/53.10 & 67.61/39.14 & 64.96/2.33 & 68.33/8.61 & 68.15/20.37 & 71.80/19.89 & 72.34/3.17 \\

MERL-ECHO~\cite{wong2025contrastive} 
& medRxiv'25
& 80.34/59.32 & 68.32/38.55 & 64.13/2.14 & 71.92/10.39 & 71.54/24.42 & 74.84/21.74 & 72.18/6.69 \\

ECG-CLIP~\cite{zhou2025diagnosis} 
& npj DM'25
& 81.71/61.14 & 70.08/41.82 & 64.28/2.15 & 73.81/10.95 & 73.06/25.61 & 75.06/21.34 & 79.23/6.90 \\

D-BETA~\cite{hung2025boosting} 
& ICML'25
& 84.32/\underline{69.67} & 72.62/\textbf{51.65} & \underline{68.20}/\underline{2.54} & 75.62/14.05 & 76.31/34.74 & 77.90/26.68 & 80.04/4.57 \\

SGERA~\cite{chen2026sgera} 
& ICML'26
& \underline{85.28}/69.56 & \underline{73.51}/49.82 & 66.84/2.44 & 76.73/15.77 & \underline{77.92}/\underline{35.58} & \underline{78.31}/\underline{27.82} & \underline{81.72}/\underline{10.74} \\

\midrule
\textbf{EchoBridge} 
& \textbf{Ours}
& \textbf{86.15}/\textbf{70.74} & \textbf{74.74}/\underline{50.98} & \textbf{69.53}/\textbf{3.48} & \textbf{78.58}/\textbf{16.49} & \textbf{79.49}/\textbf{37.94} & \textbf{79.83}/\textbf{30.08} & \textbf{83.24}/\textbf{18.89} \\
\bottomrule
\end{tabular}
\end{adjustbox}
\end{table*}

\begin{table*}[t]
\centering
\small
\setlength{\tabcolsep}{2pt}
\renewcommand{\arraystretch}{1.05}
\caption{
Finding-specific AUROC and AUPRC on PKUPH under 100\% target-domain cross-center frozen linear probing. P/N denotes the numbers of positive and negative test samples for each echocardiography-derived finding. Mod./Sev. denotes moderate-or-severe disease.
}
\label{tab:pkuph_disease_specific_auroc}
\begin{adjustbox}{max width=\textwidth,center}
\begin{tabular}{@{}lccccccccccc@{}}
\toprule
\multicolumn{1}{c}{\multirow{2}{*}{\textbf{Methods}}}
& \multirow{2}{*}{\textbf{Ref.}}
& \textbf{LVSD} & \textbf{LVDD} & \textbf{LVWMA} & \textbf{LVH} & \textbf{LAE} & \textbf{LVE} & \textbf{RAE} & \textbf{RVE} & \textbf{Mod./Sev. TR} & \textbf{Mod./Sev. MR} \\
& & {\scriptsize P/N=129/5303} & {\scriptsize P/N=2150/3282} & {\scriptsize P/N=52/5380} & {\scriptsize P/N=1053/4379} & {\scriptsize P/N=1554/3878} & {\scriptsize P/N=211/5221} & {\scriptsize P/N=48/5384} & {\scriptsize P/N=35/5397} & {\scriptsize P/N=26/5406} & {\scriptsize P/N=49/5383} \\
\midrule

\multicolumn{12}{@{}l}{\textbf{ECG-only Self-Supervised Learning}} \\
SimCLR~\cite{chen2020simple}
& ICML'20
& 84.68/19.40 & 63.84/53.78 & 90.26/13.22 & 60.60/29.82 & 59.91/37.88 & 77.42/19.48 & 72.47/2.98 & 71.51/1.24 & 69.04/1.78 & 76.37/5.52 \\

ST-MEM~\cite{na2024guiding}
& ICLR'24
& 88.78/30.41 & 65.78/\underline{57.34} & 91.57/17.18 & 63.06/31.50 & 62.14/40.83 & 81.08/26.59 & 77.48/4.78 & 79.69/9.59 & 79.03/3.94 & 79.79/6.94 \\

HeartLang~\cite{jin2025reading}
& ICLR'25
& 88.95/29.73 & 66.46/56.71 & 91.53/17.37 & 63.76/31.31 & 62.48/40.49 & 81.62/26.45 & 77.88/4.70 & 78.11/10.55 & 79.11/3.79 & 79.60/7.90 \\

\midrule
\multicolumn{12}{@{}l}{\textbf{ECG-Text Pretraining}} \\
CLIP~\cite{radford2021learning}
& ICML'21
& 87.13/27.84 & 65.35/55.96 & 90.44/16.97 & 61.94/31.03 & 60.51/39.50 & 79.08/22.86 & 74.09/3.64 & 76.45/4.88 & 72.89/2.69 & 77.42/7.13 \\

SigLIP~\cite{zhai2023sigmoid}
& ICCV'23
& 84.99/22.53 & 64.25/55.15 & 89.88/15.70 & 60.41/29.91 & 60.16/38.93 & 78.29/22.72 & 73.05/3.51 & 73.88/4.95 & 69.82/1.47 & 75.37/6.93 \\

PCME++~\cite{chun2023improved}
& ICLR'24
& 87.62/19.32 & 66.02/55.12 & 90.75/13.43 & 63.43/30.83 & 61.29/38.22 & 79.29/18.95 & 74.51/2.98 & 76.22/1.72 & 73.39/1.84 & 78.08/5.29 \\

MERL-ECHO~\cite{wong2025contrastive}
& medRxiv'25
& \underline{91.06}/25.45 & \underline{66.98}/56.46 & \underline{92.45}/15.39 & \underline{64.45}/31.16 & 62.65/39.63 & 82.58/23.68 & 77.77/4.26 & 80.00/6.86 & 79.56/3.13 & \underline{81.30}/6.38 \\

ECG-CLIP~\cite{zhou2025diagnosis}
& npj DM'25
& 90.48/33.74 & 66.81/56.47 & 92.27/18.25 & 64.23/31.15 & \underline{62.68}/39.63 & \underline{82.74}/27.45 & \underline{78.97}/4.90 & \underline{81.42}/12.88 & 79.85/3.62 & 80.75/8.11 \\

D-BETA~\cite{hung2025boosting}
& ICML'25
& 89.55/\underline{34.33} & 66.22/56.78 & 91.65/\underline{18.57} & 63.72/31.36 & 62.61/40.55 & 81.26/27.90 & 78.06/\underline{4.95} & 80.38/11.58 & 80.35/4.13 & 81.00/\underline{8.25} \\

SGERA~\cite{chen2026sgera}
& ICML'26
& 90.92/34.09 & 66.59/57.18 & 92.22/18.54 & 64.06/\underline{31.65} & 62.49/\underline{40.84} & 81.90/\underline{28.45} & 77.52/4.62 & 79.91/\underline{13.39} & \underline{80.38}/\underline{4.28} & 80.61/8.06 \\

\midrule
\textbf{EchoBridge}
& \textbf{Ours}
& \textbf{92.91}/\textbf{37.73} & \textbf{67.47}/\textbf{58.39} & \textbf{93.30}/\textbf{19.41} & \textbf{65.18}/\textbf{32.09} & \textbf{63.72}/\textbf{41.99} & \textbf{84.12}/\textbf{29.95} & \textbf{79.86}/\textbf{5.30} & \textbf{82.88}/\textbf{16.10} & \textbf{82.93}/\textbf{4.68} & \textbf{82.38}/\textbf{8.70} \\

\bottomrule
\end{tabular}
\end{adjustbox}
\end{table*}

\subsection{Definition-Only Prompt Construction for Classifier-Free Inference}

For prompt-based classifier-free inference, each echocardiography-derived finding is represented by a fixed definition-only prompt encoded by the frozen text branch as an $\ell_2$-normalized prototype. Because standardized pretraining summaries derive from structured labels, direct label-name prompts may create lexical overlap with training text. We instead describe each phenotype through target-defining functional, anatomical, hemodynamic, severity-related, and quantitative characteristics, excluding etiologies, associated abnormalities, and explicit ECG manifestations. GPT-5.5 Thinking generated initial candidates, which senior cardiologists reviewed and standardized according to predefined EchoNext-Mini label definitions. The final prompts in Table~\ref{tab:prompts} were fixed before testing and constructed independently of validation and test performance. During inference, normalized ECG representations and textual prototypes are compared by cosine similarity to obtain class-wise scores. This protocol evaluates the semantic accessibility of pretraining-seen findings without downstream classifier training.

\subsection{Downstream Classifier Training}
For all linear-probing experiments, the pretrained ECG encoder and output projection are frozen. We train a linear multi-label classifier with sigmoid outputs using binary cross-entropy with logits. Optimization uses AdamW with a learning rate of $1\times10^{-3}$, weight decay of $1\times10^{-6}$, a batch size of 128, and a maximum of 30 epochs. Gradients are restricted to the classifier parameters. All methods use the same frozen-representation protocol, classifier architecture, optimizer, hyperparameters, and validation-based model selection.

\begin{table*}[t]
\centering
\small
\setlength{\tabcolsep}{8pt}
\renewcommand{\arraystretch}{1.05}
\caption{
Finding-specific AUROC and AUPRC on SHTMU under 100\% target-domain cross-center frozen linear probing. P/N denotes the numbers of positive and negative test samples for each echocardiography-derived finding. Mod./Sev. denotes moderate-or-severe disease.
}
\label{tab:shtmu_disease_specific_auroc}
\begin{adjustbox}{max width=\textwidth,center}
\begin{tabular}{@{}lcccccccc@{}}
\toprule
\multicolumn{1}{c}{\multirow{2}{*}{\textbf{Methods}}}
& \multirow{2}{*}{\textbf{Ref.}}
& \textbf{LVSD} & \textbf{LVH} & \textbf{LAE} & \textbf{RAE} & \textbf{Mod./Sev. AR} & \textbf{Mod./Sev. TR} & \textbf{Mod./Sev. MR} \\
& & {\scriptsize P/N=369/3349} & {\scriptsize P/N=663/3055} & {\scriptsize P/N=2196/1522} & {\scriptsize P/N=28/3690} & {\scriptsize P/N=56/3662} & {\scriptsize P/N=121/3597} & {\scriptsize P/N=124/3594} \\
\midrule

\multicolumn{9}{@{}l}{\textbf{ECG-only Self-Supervised Learning}} \\
SimCLR~\cite{chen2020simple}
& ICML'20
& 80.91/48.78 & 65.13/29.08 & 59.61/61.76 & 77.99/1.09 & 59.52/1.12 & 71.92/10.14 & 74.29/12.39 \\

ST-MEM~\cite{na2024guiding}
& ICLR'24
& 88.69/53.43 & 68.44/31.20 & 62.32/66.96 & 79.35/2.30 & \underline{61.04}/1.43 & 76.52/11.63 & 80.80/14.49 \\

HeartLang~\cite{jin2025reading}
& ICLR'25
& 87.71/51.76 & 69.79/\underline{33.81} & 61.83/63.46 & 78.85/1.32 & 60.01/1.41 & 76.63/10.61 & 78.91/13.47 \\

\midrule
\multicolumn{9}{@{}l}{\textbf{ECG-Text Pretraining}} \\
CLIP~\cite{radford2021learning}
& ICML'21
& 79.88/49.91 & 64.32/30.05 & 58.78/63.26 & 77.41/2.15 & 58.82/1.16 & 71.21/9.66 & 73.35/12.09 \\

SigLIP~\cite{zhai2023sigmoid}
& ICCV'23
& 74.67/46.04 & 61.47/27.56 & 56.95/58.20 & 76.94/2.29 & 58.75/1.24 & 67.36/8.55 & 69.15/10.89 \\

PCME++~\cite{chun2023improved}
& ICLR'24
& 79.62/46.12 & 63.69/27.65 & 58.35/58.70 & 76.47/0.85 & 58.16/1.04 & 70.81/8.36 & 72.96/10.58 \\

MERL-ECHO~\cite{wong2025contrastive}
& medRxiv'25
& 84.67/49.19 & 66.65/28.90 & 60.75/61.71 & 78.80/1.84 & 60.41/1.03 & 73.95/10.55 & 77.30/12.89 \\

ECG-CLIP~\cite{zhou2025diagnosis}
& npj DM'25
& 87.91/52.08 & 68.73/30.26 & 63.09/66.02 & \underline{79.31}/\underline{2.43} & 60.43/1.66 & 78.07/11.88 & 79.90/13.96 \\

D-BETA~\cite{hung2025boosting}
& ICML'25
& 86.79/\underline{54.20} & 67.51/33.74 & 62.70/68.38 & 78.85/2.00 & 60.96/1.44 & 78.90/\underline{12.82} & 80.05/\underline{14.46} \\

SGERA~\cite{chen2026sgera}
& ICML'26
& \underline{89.01}/53.99 & \underline{69.97}/32.68 & \underline{63.33}/\underline{69.46} & 78.48/1.15 & 59.92/\underline{1.83} & \underline{79.28}/11.73 & \underline{80.60}/14.03 \\

\midrule
\textbf{EchoBridge}
& \textbf{Ours}
& \textbf{91.23}/\textbf{56.40} & \textbf{71.05}/\textbf{35.07} & \textbf{64.01}/\textbf{70.79} & \textbf{80.23}/\textbf{2.81} & \textbf{61.87}/\textbf{2.27} & \textbf{80.40}/\textbf{13.38} & \textbf{83.16}/\textbf{16.02} \\

\bottomrule
\end{tabular}
\end{adjustbox}
\end{table*}

\section{Additional Results}
\label{additional_results}
\subsection{In-Domain Finding-Specific Performance}
Table~\ref{tab:disease_specific_auroc} evaluates findings with positive prevalence ranging from approximately 25\% to below 1\%. Under 100\% frozen linear probing, EchoBridge achieves the highest AUROC for all seven findings and the highest AUPRC for six, extending its gains beyond frequent LVSD and LVH. Relative to the strongest baseline for each finding, its largest AUROC gains are 1.59 and 1.57 points for moderate-or-severe AS and TR, while its largest AUPRC gain is 8.15 points for PR. These results support improved discrimination across prevalence levels, particularly for low-prevalence valvular findings.

\subsection{Cross-Center Finding-Specific Performance}

\paragraph{\bfseries Finding-Specific Performance on PKUPH}
Table~\ref{tab:pkuph_disease_specific_auroc} reports finding-specific performance on PKUPH under 100\% target-domain cross-center frozen linear probing. EchoBridge achieves the highest AUROC and AUPRC point estimates for all ten findings. The largest gains occur for LVSD, exceeding the strongest baseline by 1.85 AUROC and 3.40 AUPRC points, followed by LVE with gains of 1.38 and 1.50 points. Improvements also cover chamber enlargement, left ventricular wall-motion abnormality, ventricular hypertrophy, and valvular findings. Moderate-or-severe TR, RVE, and RAE contain only 26, 35, and 48 positive test cases, respectively, so their AUPRC values require cautious interpretation. The consistently higher point estimates support cross-center transfer across frequent and low-prevalence findings under institutional and label-distribution shifts.

\paragraph{\bfseries Finding-Specific Performance on SHTMU}
Table~\ref{tab:shtmu_disease_specific_auroc} reports finding-specific performance on SHTMU under 100\% target-domain cross-center frozen linear probing. EchoBridge achieves the highest AUROC and AUPRC point estimates for all seven findings, with gains over the strongest baseline of 0.68--2.36 AUROC points and 0.38--2.20 AUPRC points. Improvements span frequent LAE and low-prevalence RAE and moderate-or-severe aortic, tricuspid, and mitral regurgitation, indicating linear accessibility across chamber, ventricular, and valvular abnormalities at a second independent institution. RAE and moderate-or-severe aortic regurgitation contain only 28 and 56 positive test cases, respectively, so these estimates require cautious interpretation under severe class imbalance.

\begin{table}[t]
\centering
\caption{
Comparison of global, shared, and private representations on EchoNext-Mini. Prompt-based classifier-free inference uses fixed definition-only textual prototypes, whereas 100\% frozen linear probing trains a linear multi-label classifier using the complete labeled training set.
}
\label{tab:shared_private_analysis}

\small
\setlength{\tabcolsep}{1.5pt}
\renewcommand{\arraystretch}{1.05}

\begin{adjustbox}{max width=\columnwidth}
\begin{tabular}{@{}lcccccc@{}}
\toprule
\multirow[c]{2}{*}{\textbf{Representation}}
& \multicolumn{3}{c}{\textbf{Prompt-based}}
& \multicolumn{3}{c}{\textbf{100\% Linear Probing}} \\
\cmidrule(lr){2-4}
\cmidrule(lr){5-7}
&
\textbf{AUROC}
& \textbf{AUPRC}
& \textbf{F1}
& \textbf{AUROC}
& \textbf{AUPRC}
& \textbf{F1} \\
\midrule

Global representation w/o CSPP
& 73.36
& 24.71
& 30.02
& 77.83
& 31.58
& 35.13 \\

Private representation
& --
& --
& --
& 75.67
& 26.39
& 31.30 \\

Shared representation
& \textbf{75.73}
& \textbf{26.79}
& \textbf{31.83}
& \textbf{78.79}
& \textbf{32.66}
& \textbf{35.82} \\

\bottomrule
\end{tabular}
\end{adjustbox}
\end{table}

\subsection{Shared--Private Representation Analysis}

Table~\ref{tab:shared_private_analysis} compares the global representation from the no-CSPP variant with EchoBridge's shared and private representations. The shared representation performs best under both protocols. Relative to the global representation, it improves prompt-based AUROC, AUPRC, and F1 by 2.37, 2.08, and 1.81 points, respectively, and 100\% frozen linear-probing performance by 0.96, 1.08, and 0.69 points. These gains indicate greater accessibility of echocardiography-derived finding information to textual prototypes and linear classifiers in the shared space.

The private representation remains predictive under linear probing, showing that the auxiliary branch retains task-relevant ECG information. The shared branch exceeds it by 3.12 AUROC, 6.27 AUPRC, and 4.52 F1 points, with larger AUPRC and F1 gaps indicating better positive-finding discrimination under class imbalance. Prompt-based inference is evaluated only in the shared space because the private branch is unaligned with textual prototypes. These results support the complementary roles of both branches, while characterizing private-branch information requires further modality-specific analysis.

\begin{table}[t]
\centering
\small
\caption{
Comparison with fully supervised task-specific models on EchoNext-Mini using the complete training set. The supervised baselines are trained end-to-end, whereas EchoBridge uses a frozen pretrained ECG encoder and shared projection with a linear multi-label classifier.
}
\label{tab:supervised_comparison}

\setlength{\tabcolsep}{2pt}
\renewcommand{\arraystretch}{1.05}

\begin{adjustbox}{max width=\columnwidth}
\begin{tabular}{@{}lccc@{}}
\toprule
\textbf{Methods}
& \textbf{AUROC}
& \textbf{AUPRC}
& \textbf{F1} \\
\midrule

ResNet-18 + BCE
& \CI{77.95}{77.25}{78.70}
& \CI{30.77}{29.67}{32.02}
& \CI{34.66}{33.76}{36.14} \\

ResNet-18 + ASL
& \CI{\underline{79.46}}{78.73}{80.17}
& \CI{\underline{32.75}}{31.61}{34.11}
& \CI{\underline{36.39}}{35.51}{37.94} \\

ResNet-18 + Cosine BCE
& \CI{\textbf{79.78}}{79.04}{80.48}
& \CI{\textbf{33.46}}{32.39}{34.72}
& \CI{\textbf{36.60}}{35.77}{38.01} \\

\midrule

EchoBridge
& \CI{78.79}{77.98}{79.55}
& \CI{32.66}{31.65}{33.88}
& \CI{35.82}{35.02}{37.36} \\

\bottomrule
\end{tabular}
\end{adjustbox}
\end{table}

\subsection{Comparison with Fully Supervised Models}
\label{sec:supervised_comparison}

Table~\ref{tab:supervised_comparison} compares EchoBridge's frozen representation with task-specific ECG classifiers trained end-to-end using all EchoNext-Mini training labels. All supervised baselines use the same ResNet-18 backbone and differ only in their classification heads or training objectives. ResNet-18 + BCE and ResNet-18 + ASL use linear heads optimized with binary cross-entropy and asymmetric loss, respectively. ResNet-18 + Cosine BCE uses an $\ell_2$-normalized cosine classifier with binary cross-entropy, providing a supervised reference for EchoBridge's prototype-based formulation.

Among the fully supervised models, ResNet-18 + Cosine BCE achieves the highest performance, with 79.78 AUROC, 33.46 AUPRC, and 36.60 F1. With the pretrained ECG encoder and shared projection frozen, EchoBridge obtains 78.79 AUROC, 32.66 AUPRC, and 35.82 F1 using a linear classifier trained with all probe labels. These values exceed ResNet-18 + BCE by 0.84 AUROC, 1.89 AUPRC, and 1.16 F1 points, while remaining 0.99, 0.80, and 0.78 points below the strongest fully supervised baseline, respectively.

This comparison contextualizes EchoBridge's source-domain performance under complete downstream supervision. EchoBridge provides a competitive frozen cross-modal representation while additionally supporting prompt-based classifier-free inference and source-only cross-center transfer through the same pretrained representation space.

\end{document}